\documentclass[11pt]{article}
\usepackage{enumitem}
\usepackage{amsmath}
\usepackage[final]{acl}

\usepackage{times}
\usepackage{latexsym}

\usepackage[T1]{fontenc}

\usepackage[utf8]{inputenc}

\usepackage{microtype}

\usepackage{inconsolata}

\usepackage{graphicx}

\usepackage{booktabs}
\usepackage{amssymb}

\usepackage{array}

\newcounter{notecounter}
\newcommand{\enotesoff}{\long\gdef\enote##1##2{}}
\newcommand{\enoteson}{\long\gdef\enote##1##2{{
\stepcounter{notecounter}
{\large\bf \hspace{1cm}\arabic{notecounter} $<<<$ ##1: ##2 $>>>$\hspace{1cm}}}}}
\enoteson
\enotesoff 

\usepackage{subcaption}
\usepackage{amsmath}
\usepackage{relsize}

\usepackage{booktabs}
\usepackage{multirow}
\usepackage{float}
\usepackage{xcolor}
\usepackage{rotating}
\usepackage{multirow}
\usepackage{pgfplots}
\usepackage{tikz}
\usepackage{xspace}
\usepackage{adjustbox}

\usepackage{mathabx}
\definecolor{cmzhao}{rgb}{0.1, 0.8, 0.1}

\def\secref#1{\S\ref{sec:#1}}
\def\seclabel#1{\label{sec:#1}}

\usepackage{tabularx}

\title{A Target-Centric Survey of Quantization-Aware Training}

\author{
  Jiamin Song \textsuperscript{\normalfont 1}\thanks{Equal contribution} \quad
  Mengjie Zhao \textsuperscript{\normalfont 1}\footnotemark[1] \quad
  Zijing Wang \textsuperscript{\normalfont 1} \quad
  Yongkang Liu \textsuperscript{\normalfont 1}\thanks{Corresponding authors} \quad \\
  {\bf Qian Li} \textsuperscript{\normalfont 2} \quad
  { \bf Shi Feng} \textsuperscript{\normalfont 1} \quad 
  { \bf Feiliang Ren} \textsuperscript{\normalfont 1} \quad 
  {\bf Daling Wang} \textsuperscript{\normalfont 1} \quad
  {\bf Hinrich Schütze} \textsuperscript{\normalfont 3}
  \\
  \textsuperscript{1}Northeastern University, China \quad
  \textsuperscript{2}Shandong University, China \\
  \textsuperscript{3}CIS, LMU Munich; MCML, Germany \\
}

\begin{document}
\maketitle

\begin{abstract}
The rapid development of LLMs incurs prohibitive memory footprints and intensive computational demands.
Quantization-Aware Training (QAT) techniques have emerged as a promising solution to address these challenges by explicitly \emph{simulating quantization effects during model training}, 
yielding low-bit models that achieve accuracy comparable to their full-precision counterparts. 
In this work, we provide a target-centric survey of QAT, aimed at clarifying both its theoretical foundations and its evolving implementation landscape.
We systematically review existing QAT methods through a target-centric taxonomy and synthesize cross-target differences in error characteristics, numerical formats, and strategy transferability.
We further summarize QAT evaluation paradigms and discuss challenges in optimization and deployment, outlining potential directions for future research.

\end{abstract}

\section{Introduction}
Recent years have witnessed rapid development progress of large-scale foundation models
capable of communicating across different languages, understanding diverse data modalities, and generating multimedia assets.
The rapidly growing scale of these models demands intensive memory and computational resources far exceeding the capacity of most current edge devices, sometimes even consumer-grade GPUs \citep{Nagel2021AWP,wei24surveyad,yang25surveynnq,liu2025lowbit}. 
To mitigate these resource bottlenecks, model quantization serves as a fundamental technique, compressing the size of these massive models by 
lowering the bit-width of their parameters \citep{jacob2018quantization,Nagel2021AWP,wei24surveyad}.

Model quantization methods can be broadly divided into Post-training quantization (\textbf{PTQ}) and Quantization-aware training (\textbf{QAT}; \citet{Nagel2021AWP,wei24surveyad,yang25surveynnq}). 
PTQ applies low-precision quantization to a pre-trained model without further training to reduce model size and improve inference efficiency \citep{xiao2023smoothquant,liu2024spinquant,zhang2024leanquant}. 
However, directly quantizing a pre-trained model introduces significant performance degradations due to the limited representational range of low-bit weights.
For instance, \citet{bondarenko2021understanding} identified that activations in residual connections contain structured outliers with high dynamic ranges, which are difficult to represent in low-bit formats.
This has been repeatedly observed across numerous quantization methods; researchers are then cautious about directly applying PTQ.

In contrast, \emph{QAT simulates quantization during training} using fake quantization modules, optimizing the model directly under quantized inference constraints, and often recovering the accuracy loss of PTQ, especially at ultra-low bit-widths such as INT4 \citep{jacob2018quantization, bengio2013estimating}.
Recent QAT methods further combine quantization with parameter-efficient adaptation or distillation, making them increasingly applicable to large-scale LLMs \citep{xu2023qa,chen2025efficientqat}.

There exist surveys covering QAT, however, they often discuss QAT as a component of the broader topic of neural network quantization, leaving the target-specific design principles of QAT under-explored. 
Table~\ref{tab:survey_comparison} summarizes the scope and taxonomy of representative quantization surveys and highlights how our target-centric, QAT-focused perspective differs from prior work.

\begin{table*}[t]
\centering
\small
\setlength{\tabcolsep}{5pt}
\renewcommand{\arraystretch}{1.15}

\begin{tabularx}{\textwidth}{
    >{\raggedright\arraybackslash}p{0.13\textwidth}
    >{\raggedright\arraybackslash}p{0.22\textwidth}
    >{\raggedright\arraybackslash}p{0.24\textwidth}
    >{\raggedright\arraybackslash}X
}
\toprule
\textbf{Survey} &
\textbf{Scope} &
\textbf{Taxonomy} &
\textbf{Key difference from ours} \\
\midrule

\citet{Nagel2021AWP}&
Practical neural network quantization for efficient inference. &
PTQ/QAT pipelines; hardware-oriented. &
Focuses on deployment pipelines, where QAT is one practical
component rather than the main analytical focus. \\

\addlinespace

\citet{wei24surveyad}&
General review of neural network quantization. &
Fundamentals, key techniques, deployment accuracy, and future trends. &
Provides broad coverage of quantization, with QAT discussed mainly
at a high level. \\

\addlinespace

\citet{yang25surveynnq}&
Quantization for CNN/RNN and Transformer-based models. &
Architecture-centric: traditional neural networks, LLMs, and ViTs. &
Organizes methods by model architecture, with less emphasis on
target-specific QAT mechanisms. \\

\midrule

Ours &
QAT-focused survey. &
Target-centric: weights, activations, W+A, KV cache,
and gradients. &
Centers QAT itself and explains how different quantization
targets induce distinct errors, optimization challenges, and
evaluation requirements. \\

\bottomrule
\end{tabularx}
\caption{Comparison with representative surveys on neural network
quantization.}
\label{tab:survey_comparison}
\end{table*}

In contrast, we primarily focus on QAT and organize the literature accordingly. This perspective highlights how the numerical characteristics, lifetime, reuse frequency, and deployment constraints of different quantization targets shape the design of QAT schemes, and why targeting different modules in QAT results in diverse benefits and costs.

In this work, we firstly establish theoretical foundations of QAT, focusing on the mathematical formulation of uniform quantization, the fake quantization mechanism, and gradient approximation techniques such as the Straight-Through Estimator (\secref{sec:prelim}).
Next, we organize QAT methods by quantization target, covering weights, activations, W+A, KV cache, and gradients, and synthesize their cross-target differences and strategy transferability (\secref{sec:taxonomy_target}).
Thirdly, we detail the evaluation protocols of QAT via multiple perspectives, including 
task-level effectiveness, 
structural error diagnostics, 
reasoning-aware evaluation for LLMs,
and deployment-grounded efficiency analysis.
We also propose a minimal reporting checklist, 
with detailed evaluation paradigms, metrics in Appendix~\ref{app:eval-details}, and representative quantitative evidence provided in Appendix~\ref{app:supple}
(\secref{sec:eval}).
Lastly, we discuss limitations of current QAT methods such as optimization instability and outline potential directions and opportunities for future research (\secref{sec:challenge}). 
Details of our literature search, coverage period, inclusion and exclusion criteria, and categorization procedure are provided in Appendix~\ref{app:survey_methodology}.

\section{Preliminary Knowledge}
\label{sec:prelim}
\seclabel{sec:prelim}

\subsection{Uniform Affine Quantization (UAQ)}
\label{subsec:uniform_quant}

Quantization maps high-precision (e.g., FP32) real-valued tensors to a low-precision discrete set that is more amenable to efficient integer arithmetic and reduced memory usage. 
Different bit-widths $b$ provide different trade-offs between efficiency and representational capacity. While FP32 offers high precision and a large dynamic range, lower-precision formats such as INT8 and INT4 significantly reduce memory footprint and computation cost, at the expense of potential information loss.

Let $x \in \mathbb{R}^{d_1 \times \cdots \times d_k}$ denote a real-valued tensor, e.g. a weight matrix.
UAQ is parameterized by a positive \textbf{step size} $s>0$, an integer \textbf{zero-point} $z \in \mathbb{Z}$, and an integer \textbf{representable range} $[q_{\min}, q_{\max}]$. For signed $b$-bit integers, one common choice is $q_{\min}=-2^{b-1}$ and $q_{\max}=2^{b-1}-1$. The quantization operator first maps $x$ to an integer tensor $x_{\mathrm{int}} \in \mathbb{Z}^{d_1 \times \cdots \times d_k}$:
\begin{equation}
\label{eq:quant_int}
x_{\mathrm{int}}
=
\mathrm{clamp}\!\left(
\left\lfloor \frac{x}{s} \right\rceil + z,\;
q_{\min},\; q_{\max}
\right),
\end{equation}
where $\lfloor \cdot \rceil$ denotes deterministic rounding to the nearest integer, and $\mathrm{clamp}(\cdot,q_{\min},q_{\max})$ clips values to the integer range \citep{jacob2018quantization}.

During QAT, computation is usually still performed in floating point, while low-precision effects are emulated through a quantize--dequantize (i.e., fake-quantization) operator:
\begingroup
\setlength{\abovedisplayskip}{3pt}
\setlength{\belowdisplayskip}{3pt}
\begin{equation}
\label{eq:dequant}
\hat{x} = s \cdot (x_{\mathrm{int}} - z),
\end{equation}
\endgroup
where $\hat{x}$ is the \textbf{fake-quantized} proxy of $x$ used in the forward pass. The quantization error can be written as $\epsilon = x - \hat{x}$.

UAQ is commonly instantiated in two forms. Symmetric quantization sets $z=0$, while asymmetric quantization allows $z \neq 0$.
We detail derivations of quantization parameters in Appendix~\ref{app:quant_math}, 
and Appendix~\ref{app:sub:quant_params} shows
detailed derivations of $(s,z)$ from a chosen clipping range.

\textbf{Quantization Granularity.} The scope over which $(s,z)$ are
shared defines the granularity. E.g., per-tensor quantization uses one pair for an
entire tensor while per-channel quantization is more fine-grained, targeting specific dimensions of a weight tensor \citep{Nagel2021AWP}.

\subsection{QAT and Gradient Approximation}
\label{subsec:qat_mech}

Quantization-aware training (QAT) \emph{incorporates quantization effects during training} by inserting quantization operators into the training graph \citep{jacob2018quantization,Nagel2021AWP}. 
This allows the model to effectively learn to adapt its parameters and intermediate representations to discretization effects encountered during low-precision inference.

\paragraph{Fake Quantization.}
In QAT, selected tensors are replaced by their fake-quantized counterparts during the forward pass. For a layer that computes $y=f(Wx+b)$, QAT instead computes
\begin{align}
\label{eq:qat_forward}
\begin{split}
\hat{W} &= \mathcal{Q}(W), \qquad \hat{x}=\mathcal{Q}(x), \\
y &= f(\hat{W}\hat{x}+b),
\end{split}
\end{align}
where $\mathcal{Q}(\cdot)$ denotes the quantize--dequantize operator in Eqs.~\ref{eq:quant_int}--\ref{eq:dequant}. In most QAT implementations, model parameters are not stored as integers during training. Instead, QAT maintains full-precision master weights $W$ and uses their fake-quantized versions $\hat{W}$ only in the forward pass. This design preserves small gradient updates that would otherwise vanish after discretization, while exposing the model to quantization noise during training.

\paragraph{Straight-Through Estimator.}
Eq.~\ref{eq:quant_int} contains a non-differentiable rounding operator.
The derivative of rounding is zero almost everywhere and undefined at integer thresholds:
$\frac{\partial \lfloor u \rceil}{\partial u} = 0$, \emph{a.e.}
so naive backpropagation through the quantization path would produce vanishing or ill-defined gradients, preventing effective optimization of the underlying full-precision parameters \citep{nagel2022overcoming}.
QAT often relies on surrogate gradients;
Straight-Through Estimator (STE) is the most widely used method, treating rounding as an identity operation in the backward pass \citep{bengio2013estimating}:
\begingroup
\setlength{\abovedisplayskip}{4pt}
\setlength{\belowdisplayskip}{4pt}
\begin{equation}
\label{eq:ste_basic}
\frac{\partial \mathcal{L}}{\partial x}
\approx
\frac{\partial \mathcal{L}}{\partial \hat{x}}
\cdot
\mathbb{I}\!\left(x \in [x_{\min}, x_{\max}]\right),
\end{equation}
\endgroup
where $[x_{\min},x_{\max}]$ is the clipping range and $\mathbb{I}(\cdot)$ is the indicator function. Intuitively, STE allows gradients to pass through the quantization operator as if $\hat{x}\approx x$ inside the valid range, while suppressing gradients for values that are saturated by clipping.

Though simple and efficient, STE is a coarse approximation.
Later QAT methods therefore either improve the surrogate-gradient design, learn quantizer parameters, stabilize low-bit optimization, or replace hard rounding with differentiable relaxations (cf. Appendix~\ref{app:ste_variants}).

\section{Taxonomy}
\label{sec:taxonomy_target}
\seclabel{sec:taxonomy_target}

\begin{figure*}[!t]
  \centering
  \includegraphics[width=\textwidth]{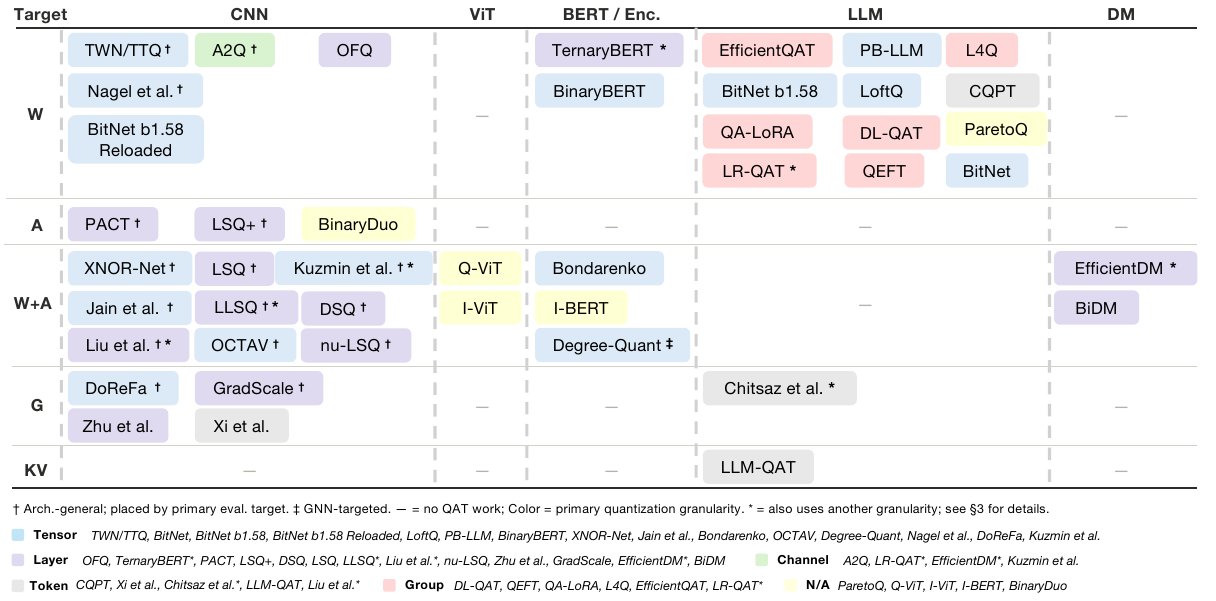}
  \caption{Taxonomy of QAT methods, organized by quantization target (rows) and model family (columns). Method name color encodes primary quantization granularity: tensor-, layer-, channel-, row-, column-, group-, token-, or instance-level quantization. Methods with multiple granularities are colored by the one their ablations center on and marked with * in the legend. † denotes architecture-general methods placed by their primary evaluation target; ‡ denotes GNN-targeted; — indicates no identified QAT work in that cell. Appendix~\ref{app:supple} Table~\ref{tab:qat_taxonomy} shows a table view.}
  \label{fig:qat_taxonomy}
\end{figure*}

Existing surveys often treat QAT as a component of general neural-network quantization or organize methods primarily by architecture \citep{Nagel2021AWP, wei24surveyad, yang25surveynnq}. 
In contrast, we organize QAT methods by quantization target, \emph{because different targets induce different sources of error and require different mitigation strategies}.
This target-centric taxonomy is therefore more diagnostic than architecture-only or bit-width-only taxonomies.
Throughout this survey, we use W\emph{n}A\emph{m} to denote n-bit weight and m-bit activation quantization (e.g., W4A4 refers to 4-bit weights and 4-bit activations).
We present an overview of our target-centric taxonomy of representative QAT frameworks in Figure~\ref{fig:qat_taxonomy}. While many methods quantize multiple targets simultaneously, we group each work under the targets where its key ideas, algorithmic designs, and ablations are centered.
Appendix~\ref{app:supple} Table~\ref{tab:qat_taxonomy} shows the full method-level categorization.

\subsection{Quantization Target: Model Weights}
\label{subsec:weight_only}
This section focuses on works applying QAT primarily to model weights.

\paragraph{Low-Bit Feasibility and Stability.} 

Early weight-QAT methods established the feasibility of low-bit training but exposed trade-offs between compression and stability \citep{zhou2016dorefa}. Binary quantization \citep{rastegari2016xnor} offers maximal compression at the cost of larger accuracy loss, while ternary methods \citep{li2016ternary} improve representational flexibility by adding a zero state.
At very low bit-widths, the mismatch between continuous weight updates and discrete quantization levels can cause weights near thresholds to repeatedly flip between adjacent values, leading to oscillation and unstable convergence \citep{nagel2022overcoming}. This problem is further amplified in low-bit ViTs by learnable scaling factors and coupled Query/Key weight quantization \citep{liu2023oscillation}. Existing methods mitigate such instability through oscillation dampening and iterative weight freezing \citep{nagel2022overcoming}, ViT-specific designs including statistical weight quantization, confidence-guided annealing, and Query-Key reparameterization \citep{liu2023oscillation}, as well as stronger supervision or staged training for ternary and binary language models \citep{zhang-etal-2020-ternarybert,bai-etal-2021-binarybert}. 
Weight QAT must also consider hardware feasibility, since low-precision deployment may suffer from accumulator overflow even when fake-quantized simulation is accurate. A2Q addresses this by incorporating accumulator-aware constraints \citep{colbert2023a2q}.

\paragraph{Selective Weight Quantization.} 
In ultra-low-bit weight QAT, (e.g., below 3 bits), the central challenge is to preserve model capacity under a highly constrained parameter space. 
Different low-bit regimes require different quantization functions and training strategies, rather than a uniform treatment of all weights\citep{liu2025paretoq}. 
Preserving a small subset of salient weights in higher precision \citep{YuanSD24}, using Hessian-based group-wise quantization, and adopting mixed-precision allocation \citep{Shen2019QBERTHB} can all reduce the degradation caused by uniform low-bit quantization.
The other line of work improves scalability by restricting optimization to quantization-related parameters, auxiliary parameters, or low-rank subspaces. 
EfficientQAT \citep{chen2025efficientqat} expands the optimization space through staged training and then updates only quantization parameters, while DL-QAT \citep{ke2024dl} formulates QAT as low-rank adaptation in a decomposed quantization space. 
LoRA-style methods further reduce QAT cost by coupling quantization with low-rank adaptation: L4Q \citep{jeon2025l4q} achieves LoRA-like training cost while producing a fully quantized final model, QA-LoRA \citep{xu2023qa} improves quantization flexibility through group-wise operators and supports lossless adapter merging, and LoftQ \citep{li2023loftq} reduces quantization error through joint backbone quantization and low-rank initialization, including experiments with non-uniform NF2/NF4 formats.
QEFT \citep{lee-etal-2024-qeft} bridges these two directions by using sensitivity-aware mixed precision and hardware-friendly reordering to preserve important weight structures while improving the efficiency of quantized fine-tuning.

\paragraph{Pre-Training Integrated QAT.} 
Recent work explores integrating QAT to pre-training: Low-bit weights are not only used to approximate a pre-trained full-precision model, but become part of the model's large-scale pre-training dynamics.
Binary Transformer pre-training shows that discrete weight parameterization can be used from the beginning of training \citep{wang2023bitnet}, while 1.58-bit ternary training provides a middle ground between binary compression and full-precision expressivity \citep{ma2024era}.
The timing of quantization is also important: native 1.58-bit training remains possible at smaller scales but may need architectural or capacity adjustments \citep{nielsen2024bitnet}, and continual quantization-aware pre-training treats the switch from high precision to 1.58-bit as a key design choice \citep{nielsen2025continual}. 

\subsection{Quantization Target: Activations}
\label{sec:a_quantization}
Activation-centric QAT quantizes intermediates produced in the forward pass, e.g, hidden states, attention outputs, and residual streams.  
Activations are input-dependent and vary across data instances, layers, tokens and timesteps. 
Activation quantization is sensitive to distribution shift \citep{jacob2018quantization,liu2024llm}, and is central to integer-only inference and memory-efficient long-sequence processing.
The Transformer architecture further challenges quantization by its
structured activation outliers: a small number of tokens dominate the dynamic range, forcing most values into poorly resolved quantization intervals \citep{bondarenko2021understanding,xiao2023smoothquant}. 
Activation QAT for diffusion models is also challenging due to temporal non-stationarity, i.e., activation statistics vary along the denoising trajectory and errors accumulate along timesteps \citep{li2023qdm,he2024efficientdm}.
At very low bit-widths, gradient mismatch between the training surrogate and the inference-time discrete operator becomes an additional bottleneck \citep{Kim2020BinaryDuoRG}.

Existing methods address the challenges from two directions:
(1) making the quantizer adaptive.
Instead of fixed symmetric ranges (Eq.~\ref{eq:quant_int}), learnable clipping thresholds, step sizes, offsets, or asymmetric ranges are optimized
so that the quantization grid 
tracks shifting activation distributions
\citep{choi2018pact,esser2020learned,bhalgat2020lsq+};
(2) reducing the mismatch between smooth training surrogates
and discrete inference operators via training–inference decoupling and smoother surrogate activations \citep{Kim2020BinaryDuoRG}.

Effective activation QAT needs to properly handle outliers, non-stationary statistics, and training–inference mismatch.
Transformer activation outliers, diffusion-model timestep shifts, and binary-network gradient mismatch each require target-specific solutions; no single quantizer design addresses all of them.

\subsection{Quantization Target: Weight + Activation}
\label{sec:wa_quantization}

Weights and activations (and other intermediates) can also be jointly quantized (\textbf{WA-QAT}) during training, making the simulated low-precision training graph closer
to the inference graph during deployment \citep{jacob2018quantization,Nagel2021AWP}. 
WA-QAT introduces coupled distortion: quantized weights reshape downstream activation distributions, activation clipping affects backward updates, and integer deployment imposes operator- and accumulator-aware constraints \citep{bondarenko2021understanding,colbert2023a2q}. 
Thus, WA-QAT is a co-design problem involving quantizer parameterization, graph-level integerization, and architecture-specific adaptation.
We identify three primary challenges in WA-QAT and survey the corresponding solutions developed to address them.

\paragraph{Range Adaptation for Coupled Quantization Noise.} 
In WA-QAT, upstream weight quantization continuously shifts downstream feature statistics, such that the activation ranges become unstable once weights and activations are discretized.
Static calibration is therefore insufficient; models need to learn the range parameters adaptively.
PACT learns activation clipping thresholds \citep{choi2018pact}, while LSQ learns quantization step sizes through STE-style gradients \citep{esser2020learned}.
LSQ+ introduces learnable offsets for asymmetric activation distributions \citep{bhalgat2020lsq+}. 
nu-LSQ extends this to non-uniform grids for heavy-tailed or multi-modal distributions \citep{gongyo2024learning}. 
FP8 QAT introduces an exponent–mantissa representation whose magnitude-dependent spacing provides additional dynamic range for jointly quantized \citep{kuzmin2022fp8}.
On the other hand, hardware-aware methods restrict these ranges to symmetric or fixed-point thresholds to guarantee integer-only compatibility \citep{zhao2020linear,jain2020trained,Sakr2022OptimalCA}. 
Overall, adaptive range strategies must therefore balance 
the need to accurately model shifting distributions against
the strict execution constraints of integer-only hardware.

\paragraph{Integer Transformers as Graph-Level Redesign.} 
Quantizing only the linear projections in a Transformer is 
in-sufficient for integer-only deployment, because numerical errors and floating-point fallbacks can still arise from operations like Softmax, GELU, and LayerNorm \citep{kim2021bert,li2023vit}. 
Integer Transformer methods therefore reformulate these operators for integer-friendly arithmetic:
I-BERT-style and I-ViT-style target integer-compatible nonlinear and normalization operations \citep{kim2021bert,li2023vit}.
Q-ViT shows that attention and feed-forward activations are the primary degradation sources in fully quantized ViTs and
introduces information rectification to mitigate this \citep{li2022qvit}. 
Together, these results show that effective Transformer WA-QAT 
should be understood as a computational-graph redesign,
not just quantizing individual tensors.

\paragraph{Heterogeneity-Aware QAT Across Structures, Inputs, and Timesteps.} 
Uniform WA-QAT assumes stable activation statistics, but this does not hold in several scenarios.
In graph neural networks (GNNs), the degree of nodes can change aggregation statistics;
degree-aware quantization avoids overly conservative global ranges \citep{tailor2021degree}. 
For vision models, instance-conditioned bit allocation assigns lower precision to easier samples and higher precision to harder ones \citep{liu2022instance}. 
In diffusion models, quantization error accumulates across denoising timesteps, motivating timestep-aware strategies to control temporal degradation \citep{zheng2024bidm,he2024efficientdm}. 
Heterogeneity-aware methods address this by adapting precision or quantization behavior to the source of variation.
Across these scenarios, adapting precision to the source of activation variability outperforms a single global configuration.

Overall, WA-QAT involves strong interaction between numerical approximation, optimization stability, and deployable integer execution. Existing solutions remain architecture-specific:
methods stabilizing ViTs are not directly transferable to GNNs,
LLMs, or diffusion models.

\subsection{Quantization Target: KV Cache}
\label{sec:kv_quantization}

As context windows and batch sizes grow, the KV cache becomes a major memory bottleneck in autoregressive decoding. 
Cached keys and values grow with sequence length and are repeatedly reused by future attention steps, making KV-cache quantization complementary to weight and/or activation quantization \citep{kwon2023efficient,liu2024kivi,hooper2024kvquant}. 
For KV-cache QAT, errors in quantized KV states may accumulate across decoding steps as cached values are reused.

The direct relevant method involving KV cache is
LLM-QAT \citep{liu2024llm}, which performs QAT by generating synthetic training data
from a full-precision teacher and jointly simulating low-bit weights, activations, and KV cache during training.
KV-cache QAT is still an emerging direction and the methods remain limited. 

Promising directions may include:
distribution-aware approaches exploiting the varying statistics of KV, RoPE, outliers \citep{liu2024kivi,hooper2024kvquant,liu2024spinquant}; 
adaptive and vector-quantized approaches compressing KV states 
via token-wise precision allocation or learned codebooks \citep{shutova2025cache,boroujeni2026don,li2025commvq}; 
retrieval-oriented approaches preserving only attention-relevant tokens \citep{he2025a2ats}.

\subsection{Quantization Target: Gradients}

Gradient QAT\footnote{In this work, we focus on methods that quantize gradients 
within the computation graph, rather than gradients synchronized in the distributed training system.}
quantizes backward pass information such as activation gradients, layer-wise errors, and parameter gradients. 
Gradient QAT directly affects optimization directions, effective step sizes, and convergence stability. 
Gradients can be quantized together with weights and activations,  with stochastic quantization, careful scaling \citep{banner2018scalable}, or partial high-precision computation \citep{wu2018wage} to ensure convergence. 
Others treat gradient QAT as an optimization-alignment problem. 
Unified INT8 training \citep{zhu2020towards} introduces direction-sensitive clipping.
Distribution-adaptive INT8 training \citep{zhao2021distribution} exploits channel-wise gradient statistics and magnitude-aware clipping to reduce range mismatch.
Uniform low-bit gradients are often unstable for LLMs.
Recent work proposes selective mechanisms like bit splitting, sampling \citep{xi2023training}, per-block scaling \citep{xi2025jetfire}, stochastic rounding, or specialized low-bit formats \citep{Chmiel2025FP4} to stabilize training.

Overall, gradient QAT is an optimization-stability problem requiring range control, direction preservation, and structure-aware selectivity. 
Future work may borrow from adjacent fields like low-rank gradient training \citep{zhao2024galore}, stochastic rounding, or low-bit floating-point formats.

\subsection{Cross-Target Analytical Synthesis}
\label{sec:cross_target}

\paragraph{Cross-target comparison.}
Table \ref{tab:cross_target_comparison} in Appendix \ref{app:supple} summarizes these differences across above targets, including their properties, dominant failure modes, error-propagation patterns, and typical adaptation. 
A key distinction is that quantization impact depends not only on local distortion, but also on target lifetime, reuse, and computational role.

\paragraph{Numerical format.}
The target distinction also changes the role of the numerical format. Table \ref{tab:int_fp_comparison} in Appendix \ref{app:supple} compares INT8/INT4 and FP8/FP4 across dimensions such as representation, error allocation, and scaling granularity.
The principal difference between INT and FP formats is error allocation \citep{kuzmin2022fp8,micikevicius2022fp8}. The advantage of FP formats is target-dependent. FP is generally more beneficial for highly dynamic activations and gradients, while relatively stationary weights often remain well suited to integer quantization \citep{kuzmin2022fp8,micikevicius2022fp8}. 

\paragraph{Transferable strategies.} Strategies for mitigating quantization errors exhibit transferability across different quantization targets.
Transferability is strongest for mechanisms that address target-agnostic numerical problems, such as local range mismatch, heterogeneous distributions, or sensitivity to precision. 
Accordingly, fine-grained scaling, adaptive range control, and mixed-precision allocation generalize relatively well across targets \citep{esser2020learned,liu2024kivi,xi2025jetfire,Chmiel2025FP4}. 
In contrast, mechanisms tied to target-specific persistence, temporal reuse, optimization semantics, or graph-level execution, such as weight-oscillation suppression \citep{nagel2022overcoming}, KV-specific quantization \citep{liu2024kivi,hooper2024kvquant}, gradient-direction preservation \citep{zhu2020towards,xi2023training}, and integer nonlinear redesign \citep{kim2021bert,li2023vit}, do not transfer directly.

\section{Evaluation}
\seclabel{sec:eval}
\label{sec:eval}
QAT evaluation is also challenging. First, the same nominal bit-width could produce diverse quantized computation graphs. E.g., a W4 evaluated with fake quantization is not directly comparable 
to a W4 model executed with integer kernels, quantized activations, KV-cache compression, and integer nonlinear operators. 
Next, task accuracy or perplexity alone could lead to omissions about activation outliers, gradient mismatch, weight oscillation, accumulator overflow, reasoning degradation, or latency regressions. 
Lastly, reported compression does not necessarily translate into deployment efficiency.
Real-world speedups depend on kernel availability, compiler support, and whether nonlinear operations fall back to the floating point \citep{jacob2018quantization,Nagel2021AWP,colbert2023a2q}.

We organize QAT evaluation around three dimensions: QAT performance, structural diagnostics, and deployment evidence. 
Detailed protocols and quantitative results are in Appendix \ref{app:eval-details} and ~\ref{app:supple}.

\subsection{Evaluation Setup and Comparability}
\label{subsec:eval-setup}
\seclabel{subsec:eval-setup}
A QAT result is interpretable only when its quantization setting is fully specified. 
We identify five reporting dimensions:
(1) quantization target (weights, activations, gradients, KV cache, optimizer states, and nonlinear operators); 
(2) quantization granularity (e.g., per-tensor, per-channel, etc);
(3) numerical format and scaling (integer/floating-point/non-uniform representations, exponent–mantissa allocation for FP  formats, scale representation and tensor/channel/group/block granularity, rounding and clipping rules, accumulator precision);
(4) training stage 
(from-scratch, post-pretraining, fine-tuning, Q-PEFT, continual quantization-aware pretraining); 
(5) hardware evidence (fake quantization, simulated integer arithmetic, supported INT/FP/MX kernels, measured latency/throughput).

QAT methods differ in factors like training budgets, quantization targets, and deployment backends. 
We synthesize evidence within comparable models and 
derive generalizable evaluation lessons rather than constructing a unified leaderboard.

Evaluation priorities also differ by target: 
weight QAT suffers from salient-weight loss, 
activation QAT from outliers and range drift, 
KV-cache QAT from accumulated attention errors, 
gradient QAT from optimization instability,
low-bit pre-training from representation drift, 
and diffusion QAT from timestep-wise error accumulation. 
Task scores alone are insufficient; 
they should be paired with structural diagnostics and deployment evidence.

\subsection{Empirical Findings}
\label{subsec:eval-synthesis}

\textbf{Finding 1: INT8 QAT is largely mature, but deployment evidence still matters.}
For CNNs and Transformer encoders, 
INT8 QAT is near-lossless:
Unified INT8 training reports only a $0.26$ Top-1 drop on ResNet-50 \citep{zhu2020towards}, and I-BERT achieves essentially lossless W8A8 inference for RoBERTa \citep{kim2021bert}. Similar trends appear in I-ViT, where W8A8 DeiT-S slightly exceeds the full-precision baseline \citep{li2023vit}. 
The remaining challenge is deployment fidelity: 
whether the fake-quantized training graph matches the real integer execution graph \citep{jacob2018quantization,kim2021bert,li2023vit,colbert2023a2q}.

\textbf{Finding 2: W4 QAT is practical, but failure modes are architecture-dependent.}
On residual CNNs, W4A4 QAT can match or slightly exceed full precision shown by LLSQ and LSQ-style methods \citep{zhao2020linear,esser2020learned,bhalgat2020lsq+}. 
Group-wise W4 methods like LR-QAT \citep{bondarenko2024low} and
LoftQ \citep{li2023loftq}
slightly affect LLM perplexity. 
However,  failure modes vary: compact or depthwise architectures like
MobileNetV2 are sensitive to activation clipping \citep{Sakr2022OptimalCA}, ViTs degrade through attention and nonlinear operators
\citep{li2022qvit,kim2021bert,li2023vit}, 
and for LLMs, weight-only W4 is far easier than joint WA-QAT or KV-cache QAT 
\citep{liu2024llm}.

\textbf{Finding 3: W2 and binary QAT are qualitatively different from W4.}
Degradations at sub-3-bit are sharp across model families: 
Q-ViT sees -7.80 Top-1 on DeiT-S at W2A2 \citep{li2022qvit}, 
EfficientQAT W2g64 reports -4.72 zero-shot ability and +1.39 perplexity, 
on LLaMA-2-7B \citep{chen2025efficientqat}.
These results reflect a new regime dominated by STE-induced gradient mismatch, weight oscillation, salient-weight loss, and error accumulation  \citep{yin2019understanding,nagel2022overcoming,liu2023oscillation,zheng2024bidm}.

\textbf{Finding 4: Perplexity is insufficient for LLM QAT.}
Perplexity measures local next-token fidelity but does not capture long-horizon reasoning. W4 methods may report small perplexity changes while reasoning benchmarks like GSM8K degrade substantially \citep{YuanSD24}, and this gap widens at W2/NF2 \citep{li2023loftq}.

Conversely, Q-PEFT methods may inflate MMLU or commonsense scores via adaptation gains unrelated to quantization quality \citep{xu2023qa,ke2024dl,lee-etal-2024-qeft,jeon2025l4q}.

\textbf{Finding 5: Compression does not imply acceleration.}
Real-world speedup depends on kernel availability, compiler support, and whether nonlinear operators fall back to floating point. 
I-ViT achieves a 3.72–4.11$\times$ inference speedup over FP32 baselines through integer-only execution of the entire computational graph, while pipelines retaining floating-point nonlinear operators show smaller speedups in the same hardware evaluation \citep{li2023vit}.
QAT evaluations should therefore report measured latency or throughput, not only nominal compression.

\subsection{Structural, Reasoning, and Deployment Diagnostics}
\label{subsec:eval-diagnostics}

Beyond task scores, structural diagnostics are needed to locate where quantization error enters and how it propagates. 
Basic structural diagnostic metrics include normalized MSE, cosine similarity, clipping ratio, saturation ratio, and layer-wise error profiles \citep{choi2018pact,esser2020learned,bhalgat2020lsq+,jain2020trained,zhao2020linear}. 
For Transformers, these should be paired with block-output drift and KL divergence between full-precision and quantized logits, since local perturbations can be amplified by attention, normalization and other nonlinear operators 
\citep{Hinton2015DistillingTK,kim2021bert,liu2024llm}.

For LLM QAT, reasoning should be evaluated separately from perplexity. Quantization noise can accumulate across layers and decoding steps, leading to reasoning degradation while perplexity changes are small.
A minimal protocol should cover mathematical reasoning (GSM8K/MATH; \citet{cobbe2021training,hendrycks2021math}), multi-step knowledge reasoning (BBH/MMLU; \citet{suzgun2023challenging,hendrycks2020mmlu}) and code generation (HumanEval/MBPP; \citet{chen2021evaluating,austin2021program}).
Fixed prompting and decoding configurations should be reported to ensure comparability.

Efficiency evaluation should distinguish model footprint, 
training cost, and realized inference speed. 
Model footprint may include quantized weights, scales, zero-points, preserved full-precision tensors, adapters, and KV-cache storage when applicable \citep{li2023loftq,chen2025efficientqat}.
Training and inference efficiency should be reported separately, as QAT often maintains full-precision master weights, optimizer states and other information during training. 
For LLMs, prefill latency and decoding throughput should also be distinguished \citep{liu2024llm}.

\subsection{Minimal Reporting Checklist}
\label{subsec:minimal-eval}

QAT evaluation should identify where quantization works, why quantization fails, and whether compression translates into deployable efficiency.
We propose a minimal reporting checklist for future QAT-related work:
Firstly, authors should specify the full quantization configuration (\secref{subsec:eval-setup}) and evaluate at least two bit-widths with matched full-precision baselines and absolute deltas.
Secondly, results should be accompanied by at least
one structural diagnostic such as layer-wise distortion, gradient mismatch or overflow rate \citep{nagel2022overcoming}, and at least one training- or deployment-efficiency measurement 
\citep{bondarenko2024low}.
Lastly, for LLMs, perplexity alone is insufficient; at least one reasoning-intensive benchmark should be included \citep{li2023loftq,YuanSD24}.
This checklist improves reproducibility while capturing the central lesson of existing evidence:
QAT performance depends on the interaction among bit-width, quantization target, architecture, and deployment backend.
To demonstrate its practical use, Appendix~\ref{app:case_study} applies this checklist to LLM-QAT \citep{liu2024llm} as a representative worked example.

\section{Challenges and Future Directions}
\label{sec:challenge}
\seclabel{sec:challenge}

\textbf{Optimization Instability.}
STE-based surrogate gradients do not reflect the true
derivatives of the quantized objective, introducing gradient bias and mismatch \cite{Sakr2022OptimalCA}, degrading optimization stability and convergence.

Near quantization thresholds, small weight updates cause abrupt value changes, leading to oscillation \cite{liu2023oscillation}.
In ultra-low-bit regimes, coarse quantization can trap weights within a single quantization bin \cite{lee2021network}, preventing effective optimization.
Better alignment between backward gradients and the discrete objective remains an open problem.

\textbf{Activation Outliers.}
Strong input-dependent outliers in attention modules and residual streams can skew quantization ranges and degrade accuracy, especially for Transformers and LLMs \cite{bondarenko2021understanding}. 
Under aggressive joint WA-QAT, this noise accumulates across
layers and interacts unfavorably with LayerNorm and Softmax. 
Current methods like knowledge distillation and rotation-based transformations add significant overhead \cite{zhang-etal-2020-ternarybert,liu2024spinquant}.
Future work may explore quantization-native architectures or
training objectives suppressing outliers, reducing reliance on
post-hoc corrections.

\textbf{Simulation-to-Deployment Gap.}
Simulated fake quantization in FP32 does not capture low-level hardware constraints such as accumulator overflow, rounding modes, and operator fusion \cite{jacob2018quantization,colbert2023a2q}.
Models converged under QAT simulation may degrade or fail to execute efficiently on integer-only accelerators \cite{kim2021bert,li2023vit}.
Future work should aim to close this gap by incorporating 
hardware-aware constraints during QAT training.

\textbf{High Training Overhead.}
QAT requires full-precision master weights, 
optimizer states, and often distillation teachers,
making training memory and compute costs prohibitive 
for billion-parameter models \cite{liu2024llm,chen2025efficientqat}.
This creates an inherent tension between compression objectives and the cost of the compression process itself \cite{dremov2025compute,huang2024robust}.
Reducing QAT overhead without sacrificing quantization quality could be a promising direction.

\textbf{Homogenization.}
Uniform quantizers, fixed bit-width allocation, and layer-agnostic design choice can constrain expressive capacity, causing architecturally diverse models to converge to similar weight distributions and activation patterns after quantization \cite{Nagel2021AWP}.
This effect intensifies with increased model-scale, suppressing fine-grained features and limiting downstream adaptability to various tasks \cite{Yu20248bitTI}.
Balancing compression efficiency and representational diversity remains an open research challenge.

\textbf{Format–Target Co-Design.}
Emerging FP8/FP4 and microscaling formats motivate target-aware QAT
\citep{micikevicius2022fp8, rouhani2023microscaling, Chmiel2025FP4}.
Future work should jointly optimize numerical format, scaling, and rounding across W/A/KV/G, while accounting for their distinct precision and dynamic-range requirements
\citep{Chmiel2025FP4}.

\section{Conclusion}
We review quantization-aware training methods via a target-centric taxonomy covering weights, activations, KV cache, and gradients, 
and synthesize how target-specific properties shape error propagation, numerical-format choices, and strategy transferability.
We also review QAT evaluations and propose a minimal reporting checklist for future research.
We identify persistent challenges like optimization instability, 
activation outliers, simulation-to-deployment gap, high training overhead, model homogenization, and outline directions for future research.

\section*{Limitations}
This paper focuses primarily on quantization-aware training, emphasizing quantization methods, evaluations, outlining challenges and future directions. Some deployment-oriented systems and hardware-specific optimization techniques may be underrepresented.
Since existing studies differ in setup, the quantitative evidence in this survey should be interpreted as general trends rather than strict head-to-head comparisons.
Direct QAT literature remains sparse for emerging targets such as KV-cache quantization, reflecting the current maturity of these areas rather than omissions in our coverage.
Although we cover representative works, newly released methods, benchmarks, and deployment backends may not be fully reflected in this survey.
We present supplementary mathematical details of quantization parameters in Appendix~\ref{app:quant_math}, surrogate-gradient variants in Appendix~\ref{app:ste_variants}, quantitative evaluations in Appendix~\ref{app:eval-details} and \ref{app:supple}.

\section*{Acknowledgments}

The work is supported by National Science Foundation for Young Scientists of China (No.62502081, 62402293), the National Natural Science Foundation of China (No.62576085, 62272092), and the Fundamental Research Funds for the Central Universities under Grants (N2523011).

\bibliography{custom}

\appendix
\label{sec:appendix}

\appendix

\section{Survey Methodology} 
\label{app:survey_methodology} 

\paragraph{Search scope and coverage.} 
we searched using terms including QAT, quantization-aware training, low-bit training, fake quantization, integer-only training, LLM QAT, KV-cache quantization, and gradient quantization, supplemented by backward and forward citation tracing from prior surveys and representative studies. The coverage spans early low-bit QAT work through the search cutoff (May 2026), with an emphasis on work published since 2020. We prioritize peer-reviewed studies while clearly identifying recent preprints that are influential or directly relevant to emerging directions. We searched major machine-learning, NLP, and computer-vision venues, including ACL-family conferences, NeurIPS, ICML, ICLR, CVPR, ICCV, and ECCV, together with arXiv for recent emerging work.

\paragraph{Inclusion and exclusion criteria.} 
We include methods explicitly simulating quantization during training, optimize quantization-related parameters, or construct a low-bit training graph. Pure PTQ, communication-only compression, and hardware implementations without a direct QAT component are excluded from the main taxonomy. Methods involving multiple targets are categorized according to their central technical design, primary ablations, and main evaluation target.

\section{Mathematical Details of Quantization Parameters}
\label{app:quant_math}

This appendix provides the mathematical details for deriving uniform quantization parameters and for computing surrogate gradients of learnable step sizes. The main text introduces these concepts at a high level in Section~\ref{sec:prelim}; here we provide the corresponding derivations.

\subsection{Deriving Uniform Quantization Parameters}
\label{app:sub:quant_params}

A common choice in uniform affine quantization is to derive the step size $s$ and zero-point $z$ from a chosen clipping range $[x_{\min},x_{\max}]$. For asymmetric quantization, the real-valued interval $[x_{\min},x_{\max}]$ is mapped to the integer interval $[q_{\min},q_{\max}]$. The step size is given by
\begin{equation}
\label{eq:asym_scale}
s =
\frac{x_{\max}-x_{\min}}{q_{\max}-q_{\min}}.
\end{equation}
The zero-point is chosen so that the real value zero is represented as accurately as possible in the integer domain:
\begin{equation}
\label{eq:asym_zero_point}
z =
\left\lfloor
q_{\min} - \frac{x_{\min}}{s}
\right\rceil.
\end{equation}
In practice, $z$ is further clipped to the valid integer range $[q_{\min},q_{\max}]$ if necessary. The clipping range $[x_{\min},x_{\max}]$ can be estimated from calibration statistics, optimized by minimizing local reconstruction error, or learned jointly with model parameters.

For symmetric quantization, the zero-point is typically fixed to $z=0$, and the clipping range is chosen to be symmetric around zero. Let
\begin{equation}
\label{eq:sym_alpha}
\alpha = \max(|x_{\min}|, |x_{\max}|).
\end{equation}
A common symmetric step size is
\begin{equation}
\label{eq:sym_scale}
s = \frac{\alpha}{q_{\max}}.
\end{equation}
The quantize--dequantize operation then simplifies to
\begin{equation}
\label{eq:sym_fake_quant}
\hat{x}
=
s \cdot
\mathrm{clamp}
\left(
\left\lfloor \frac{x}{s} \right\rceil,\;
q_{\min},\;
q_{\max}
\right).
\end{equation}
Symmetric quantization is commonly used for weight tensors because it simplifies integer arithmetic, while asymmetric quantization is often useful for activation tensors with shifted or non-negative distributions.

\subsection{Gradient Derivation for Learnable Step Size}
\label{app:sub:lsq_grad}

Methods such as LSQ treat the quantization step size $s$ as a learnable parameter \citep{esser2020learned}. Consider the symmetric quantization case with $z=0$. The fake-quantized output can be written as
\begin{equation}
\label{eq:lsq_fake_quant}
\hat{x}
=
s \cdot
\mathrm{clamp}
\left(
\left\lfloor \frac{x}{s} \right\rceil,\;
q_{\min},\;
q_{\max}
\right).
\end{equation}
Let $v=x/s$. Under the STE approximation, the derivative of $\hat{x}$ with respect to the step size $s$ is approximated piecewise as
\begin{equation}
\label{eq:lsq_step_grad}
\frac{\partial \hat{x}}{\partial s}
=
\begin{cases}
q_{\min}, & v \le q_{\min}, \\[3pt]
\left\lfloor v \right\rceil - v, & q_{\min} < v < q_{\max}, \\[3pt]
q_{\max}, & v \ge q_{\max}.
\end{cases}
\end{equation}
Equivalently, substituting $v=x/s$ gives
\begin{equation}
\label{eq:lsq_step_grad_expanded}
\frac{\partial \hat{x}}{\partial s}
=
\begin{cases}
q_{\min}, & \frac{x}{s} \le q_{\min}, \\[3pt]
\left\lfloor \frac{x}{s} \right\rceil - \frac{x}{s}, & q_{\min} < \frac{x}{s} < q_{\max}, \\[3pt]
q_{\max}, & \frac{x}{s} \ge q_{\max}.
\end{cases}
\end{equation}
This gradient allows the quantization grid to be adjusted according to the task loss. Intuitively, the learnable step size controls the trade-off between clipping-induced distortion and rounding-induced error. LSQ further applies gradient scaling to stabilize the magnitude of step-size updates relative to weight updates \citep{esser2020learned}.

\section{Surrogate-Gradient Variants and Stabilization Techniques}
\label{app:ste_variants}

Section~\ref{sec:prelim} introduces STE as the basic mechanism that enables backpropagation through non-differentiable quantization operators. This appendix provides a more detailed discussion of common strategies for improving or bypassing vanilla STE, as well as complementary stabilization techniques used in QAT.

\subsection{Improving and Bypassing STE}
\label{app:sub:ste_improvements}

Surrogate-gradient design is a core component of QAT because it determines whether the quantized objective can be optimized stably. Existing methods improve or bypass vanilla STE along several directions, including clipped gradients, learnable quantization parameters, gradient scaling, oscillation-aware optimization, generalized activation functions, and differentiable soft quantization.

\paragraph{Clipped and masked STE.}
A basic refinement is to make the backward approximation more consistent with the clipping behavior in the forward pass. The clipped STE in Eq.~\ref{eq:ste_basic} masks gradients outside the valid quantization range, reducing the mismatch between forward saturation and backward propagation. This form is widely used in practical QAT pipelines \citep{jacob2018quantization,esser2020learned}. However, it also introduces dead zones: once weights or activations are saturated, their gradients may vanish, making it difficult for them to re-enter the quantization range. This motivates methods that learn or adapt clipping ranges during training.

\paragraph{Learnable quantization parameters.}
Another important direction extends surrogate gradients from quantized tensors to quantizer parameters themselves. Instead of using fixed scales, offsets, or clipping thresholds, these methods optimize the quantization grid jointly with model parameters. PACT learns activation clipping thresholds to balance clipping-induced distortion and rounding-induced error \citep{choi2018pact}. LSQ treats the quantization step size as a learnable parameter and updates it through an STE-style surrogate gradient with proper gradient scaling \citep{esser2020learned}. LSQ+ further introduces learnable offsets to better fit asymmetric activation distributions \citep{bhalgat2020lsq+}. Similarly, TQT optimizes quantization thresholds for fixed-point inference \citep{jain2020trained}. These methods improve STE-based QAT not by changing the rounding operator itself, but by making the quantization range and resolution task-adaptive. The detailed piecewise gradient derivation for learnable step sizes in LSQ-style methods is provided in Appendix~\ref{app:sub:lsq_grad}.

\paragraph{Gradient scaling and magnitude-aware differentiation.}
Since STE can distort not only the existence of gradients but also their scale, several methods modify the surrogate update to better preserve optimization dynamics. LSQ introduces gradient scaling for the learnable step size so that its update magnitude is comparable to that of model weights \citep{esser2020learned}. OCTAV further studies optimal clipping and magnitude-aware differentiation, showing that better gradient treatment for quantization parameters can improve low-bit QAT stability \citep{Sakr2022OptimalCA}. In low-precision training, related techniques such as GradScale use dynamic scaling and overflow feedback to stabilize quantized gradients \citep{sun2020ultra}. These methods suggest that effective STE design should account for gradient magnitude, saturation behavior, and the interaction between quantized forward computation and backward updates.

\paragraph{Oscillation-aware optimization.}
A further limitation of vanilla STE is threshold oscillation. Because the forward quantizer is piecewise constant, small full-precision updates around a quantization threshold can cause the quantized value to flip repeatedly across training steps. Such oscillation can prevent convergence and is especially harmful in low-bit regimes. Oscillation-aware QAT methods explicitly analyze this phenomenon and modify the update or quantization procedure to reduce unstable threshold crossings \citep{nagel2022overcoming,liu2023oscillation}. These methods do not necessarily replace STE, but they make STE-based optimization more stable by accounting for the discrete structure of the quantized parameter space.

\paragraph{Generalized activation functions.}
In ultra-low-bit regimes, especially binarization, the extreme nonlinearity of the sign function makes STE gradients highly unreliable \citep{hubara2016binarized}. Instead of only modifying the backward pass, some methods change the forward parameterization so that the network becomes easier to optimize under STE. ReActNet introduces learnable activation components such as RSign and RPReLU to reshape activation distributions and improve the compatibility between binary forward computation and surrogate-gradient training \citep{liu2020reactnet}. Such generalized activation functions reduce gradient mismatch by increasing the expressive flexibility of binarized networks, thereby narrowing the performance gap with full-precision models.

\paragraph{Differentiable soft quantization.}
A more radical strategy is to replace hard rounding with differentiable relaxations during training. Differentiable Soft Quantization (DSQ) formulates quantization as a smooth parametric function and uses an annealing process to gradually approach the discrete target grid \citep{gong2019differentiable}. This provides well-defined gradients during most of training and alleviates the non-differentiability of rounding. However, soft quantization introduces additional hyperparameters and may create a relaxation-to-deployment gap if the smooth training objective does not perfectly match the final hard quantizer. This trade-off is particularly relevant for noise-sensitive models such as diffusion models, where smooth optimization can help prevent error accumulation across denoising steps \citep{li2023qdiffusion}.

\paragraph{Practical trade-offs.}
Overall, STE improvement in QAT can be understood as a spectrum. Clipped or masked STE is simple and efficient, making it suitable for moderate bit-widths such as INT8 or INT4. Learnable scales, offsets, and clipping thresholds improve adaptability to non-stationary tensor distributions, especially for activation quantization. Gradient scaling, magnitude-aware differentiation, and oscillation-aware optimization improve stability in more aggressive low-bit regimes. Generalized activation functions are particularly important for binarized networks, where the sign function makes vanilla STE unreliable. Differentiable soft quantization reduces reliance on hard STE by providing smoother gradients, but it increases algorithmic complexity and may introduce a mismatch between relaxed training and discrete inference. Therefore, the choice of STE variant or alternative should depend on bit-width, quantization target, model architecture, and deployment constraints.

\subsection{Additional Stabilization Techniques}
\label{app:sub:qat_stabilization}

Beyond surrogate-gradient design, QAT often incorporates additional stabilization strategies. Knowledge distillation can provide smoother supervision from a full-precision teacher, which is especially useful when aggressive quantization causes large prediction shifts \citep{Hinton2015DistillingTK,kim-etal-2022-understanding}. Regularization terms can also be used to constrain quantized representations or reduce sensitivity to rounding noise. These techniques are complementary to STE improvements: surrogate gradients make optimization possible, while distillation and regularization help stabilize the training trajectory under severe low-precision constraints.

\section{Detailed Evaluation Protocols}
\label{app:eval-details}

This appendix provides detailed benchmark protocols and extended diagnostic recommendations. These details support the synthesis in Section~\ref{sec:eval}, but are kept in the appendix because protocols are comparable mainly within model families rather than across the entire QAT literature.

\subsection{Benchmarks and Metrics}
\label{subsec:bench-eval}

Table~\ref{tab:eval_protocols} summarizes the common evaluation protocols used in QAT studies. The protocol maturity varies substantially across model families. CNN and BERT-style encoder quantization have relatively standardized benchmarks, while ViT, diffusion, and LLM QAT require more diagnostic evaluation because quantization errors are often localized in activation outliers \citep{bondarenko2021understanding}, attention maps \citep{li2022qvit}, nonlinear operators \citep{kim2021bert,li2023vit}, KV cache \citep{liu2024llm,bondarenko2024low}, or denoising timesteps \citep{he2024efficientdm,zheng2024bidm}.

\begin{table*}[!t]
\centering
\footnotesize
\setlength{\tabcolsep}{2.2pt}
\renewcommand{\arraystretch}{1.12}
\resizebox{\textwidth}{!}{%
\begin{tabular}{@{}p{1.35cm}p{3.0cm}p{2.15cm}p{2.15cm}p{1.55cm}p{3.2cm}@{}}
\toprule
\textbf{Family} &
\textbf{Representative works} &
\textbf{Benchmarks} &
\textbf{Metrics} &
\textbf{Bits} &
\textbf{Evaluation purpose} \\
\midrule
CNNs &
DoReFa-Net, PACT, LSQ, LLSQ, A2Q, Unified INT8 &
ImageNet, CIFAR &
Top-1/Top-5, overflow, latency &
W8A8, W4A4, W2A2, W1A1 &
Evaluate low-bit convolution, activation clipping, and integer-training feasibility. \\
\midrule
ViT &
Q-ViT, I-ViT, OFQ, INT4 training &
ImageNet / ImageNet-1K &
Top-1, attention distortion, speedup &
W8A8, W4A4, W2A2 &
Measure sensitivity of attention, Softmax, GELU, and LayerNorm. \\
\midrule
BERT &
TernaryBERT, BinaryBERT, I-BERT &
GLUE, SQuAD v1.1/v2.0 &
Accuracy, F1, EM, compression &
W8A8, W2, W1 &
Test whether encoder representations survive ultra-low-bit compression. \\
\midrule
LLMs &
LLM-QAT, LR-QAT, QA-LoRA, LoftQ, EfficientQAT, PB-LLM &
WikiText-2, C4, MMLU, ARC, GSM8K, HumanEval &
PPL, zero-shot accuracy, reasoning, pass@1 &
W4, W3, W2, NF4, NF2 &
Assess language modeling, knowledge, instruction following, and reasoning. \\
\midrule
Native low-bit LMs &
BitNet, BitNet b1.58, CQPT, ParetoQ &
Pretraining corpora, zero-shot suites &
Loss, PPL, downstream accuracy &
W1, W1.58, W2--W4 &
Study whether quantization should be integrated into pretraining. \\
\midrule
Diffusion models &
EfficientDM, BiDM &
ImageNet, LSUN, COCO-style generation &
FID, IS, CLIP score, sampling cost &
W4A8, W4A4, W2A8, W1A8 &
Evaluate error accumulation along denoising trajectories. \\
\midrule
GNNs &
Degree-Quant &
Citation and inductive graph datasets &
Accuracy, generalization to unseen graphs, CPU speedup &
INT8, INT4 &
Evaluate degree-dependent quantization error and robustness to graph-structure heterogeneity. \\
\bottomrule
\end{tabular}%
}
\caption{Common evaluation protocols for QAT. Benchmarks include ImageNet \citep{deng2009imagenet}, GLUE \citep{wang2018glue}, SQuAD \citep{rajpurkar2016squad}, WikiText-2 \citep{merity2016pointer},  \citep{raffel2020exploring}, MMLU \citep{hendrycks2020mmlu}, GSM8K \citep{cobbe2021training}, HumanEval \citep{chen2021evaluating}, and FID \citep{heusel2017gans}.}
\label{tab:eval_protocols}
\end{table*}

Tables~\ref{tab:llm-coverage}--\ref{tab:gnn-coverage} provide a more detailed view of benchmark and evaluation-dimension coverage across representative QAT methods. Specifically, Table~\ref{tab:llm-coverage} summarizes the coverage of language modeling, commonsense, knowledge, reasoning, and code benchmarks for
LLM QAT methods; Tables~\ref{tab:cnn-coverage}, \ref{tab:vit-coverage}, and \ref{tab:bert-coverage} report evaluation coverage for CNNs, ViTs, and BERT-style encoders; and Tables~\ref{tab:diffusion-coverage} and \ref{tab:gnn-coverage} summarize the corresponding coverage for diffusion models and GNNs. 
For these tables, \checkmark indicates that the benchmark or evaluation dimension is reported in the main evaluation setting of the corresponding method, $\triangle$ indicates auxiliary, partial, simulated, theoretical, or otherwise less directly comparable evidence, and ``--'' indicates not reported. Coverage rate counts both \checkmark and $\triangle$ as coverage, while $\triangle$ flags comparability limitations. These coverage rates should be interpreted as descriptive summaries rather than leaderboard-style comparisons, because papers differ in backbones, training budgets, benchmark choices, kernels, accelerators, deployment assumptions, and reporting protocols.
These tables show that QAT evaluation remains uneven across model families: mature settings such as CNNs and encoder models have relatively standardized benchmarks, whereas LLMs, diffusion models, and GNNs still exhibit substantial gaps in reasoning, deployment, temporal, or structure-specific evaluation.

\begin{table*}[!t]
\centering
\scriptsize
\setlength{\tabcolsep}{2.0pt}
\renewcommand{\arraystretch}{1.08}
\resizebox{\textwidth}{!}{%
\begin{tabular}{@{}llccccccc@{}}
\toprule
\multirow{2}{*}{\textbf{Method}} &
\multirow{2}{*}{\textbf{Backbone}} &
\multicolumn{2}{c}{\textbf{Lang. Modeling}} &
\multicolumn{2}{c}{\textbf{Commonsense}} &
\multicolumn{3}{c}{\textbf{Higher-order Reasoning}} \\
\cmidrule(lr){3-4}
\cmidrule(lr){5-6}
\cmidrule(lr){7-9}
& & WT2 & C4 & Zero-shot & MMLU & GSM8K & MATH/BBH & HumanEval \\
\midrule
\multicolumn{9}{@{}l}{\textit{Base model compression (W-only / W+A, no fine-tuning)}} \\
\midrule
LLM-QAT      & LLaMA-1 7\&13\&30B        & \checkmark & \checkmark & \checkmark & \checkmark      & --        & -- & -- \\
LR-QAT       & LLaMA-1/2/3, Mistral   & \checkmark & --         & \checkmark & --              & --        & -- & -- \\
EfficientQAT & LLaMA-2/3 7\&13\&30B       & \checkmark & \checkmark & \checkmark & $\vartriangle$  & --        & -- & -- \\
DL-QAT       & LLaMA-1/2 7\&13B         & \checkmark & --         & \checkmark & $\vartriangle$  & --        & -- & -- \\
PB-LLM       & LLaMA-1 7B, OPT 1.3B             & \checkmark & \checkmark & \checkmark & --              & --        & -- & -- \\
ParetoQ      & LLaMA-3, MobileLLM       & \checkmark & --         & \checkmark & --              & --        & -- & -- \\
\midrule
\multicolumn{9}{@{}l}{\textit{Q-PEFT (quantization-aware parameter-efficient fine-tuning)}} \\
\midrule
QA-LoRA      & LLaMA-1/2 7\&13B              & --         & --         & \checkmark & \checkmark      & --        & -- & -- \\
LoftQ        & LLaMA-2 7\&13B         & \checkmark & --         & --         & --              & \checkmark & -- & -- \\
L4Q          & LLaMA, OpenLLaMA, Mistral       & --         & --         & \checkmark & \checkmark      & --        & -- & -- \\
QEFT         & LLaMA-2 7\&13\&70B               & $\vartriangle$ & --      & \checkmark & \checkmark      & \checkmark & -- & -- \\
\midrule
\multicolumn{9}{@{}l}{\textit{Native low-bit pre-training}} \\
\midrule
BitNet b1.58 & BitNet-LM ($\geq$2B)   & \checkmark         & \checkmark     & \checkmark & --      & -- & -- & -- \\
CQPT         & BitNet-LM              & --         & --         & \checkmark & --              & --        & -- & -- \\
\midrule
\textbf{Coverage rate} & \textbf{(out of 12)}
& \textbf{9/12} & \textbf{4/12} & \textbf{11/12} & \textbf{6/12}
& \textbf{2/12} & \textbf{0/12} & \textbf{0/12} \\
\bottomrule
\end{tabular}%
}
\caption{Benchmark coverage of representative LLM QAT methods. Zero-shot task sets differ across papers, so aggregate averages are not strictly comparable. LoftQ evaluates GSM8K in a fine-tuning / natural-language-generation setting. BitNet b1.58 results refer to the 2B4T technical report.}
\label{tab:llm-coverage}
\end{table*}

\begin{table*}[!t]
\centering
\scriptsize
\setlength{\tabcolsep}{2.0pt}
\renewcommand{\arraystretch}{1.08}
\resizebox{\textwidth}{!}{%
\begin{tabular}{@{}llccccccc@{}}
\toprule
\multirow{2}{*}{\textbf{Method}} &
\multirow{2}{*}{\textbf{Backbone}} &
\multicolumn{2}{c}{\textbf{Image Classification}} &
\multicolumn{2}{c}{\textbf{Architecture Coverage}} &
\multicolumn{3}{c}{\textbf{Efficiency / Deployment Evidence}} \\
\cmidrule(lr){3-4}
\cmidrule(lr){5-6}
\cmidrule(lr){7-9}
& & ImageNet & CIFAR/SVHN & ResNet/AlexNet & Compact CNNs & Low-bit Grad. & Latency/Speedup & Hardware-aware \\
\midrule
\multicolumn{9}{@{}l}{\textit{Early low-bit CNN QAT}} \\
\midrule
DoReFa-Net      & AlexNet / CNN              & \checkmark & \checkmark & \checkmark & --          & \checkmark & $\vartriangle$ & $\vartriangle$ \\
PACT            & AlexNet / ResNet           & \checkmark & \checkmark & \checkmark & --          & --         & $\vartriangle$ & $\vartriangle$ \\
DSQ             & ResNet / general CNN       & \checkmark & \checkmark & \checkmark & \checkmark          & --         & \checkmark             & \checkmark \\
LSQ             & ResNet                     & \checkmark & --         & \checkmark & $\vartriangle$          & --         & --             & -- \\
LLSQ            & ResNet / MobileNetV2       & \checkmark & \checkmark & \checkmark & \checkmark  & --         & $\vartriangle$             & $\vartriangle$ \\
\midrule
\multicolumn{9}{@{}l}{\textit{Learnable thresholds, scaling, and integer-oriented CNN QAT}} \\
\midrule
TQT & VGG/ Inception/ ResNet/ MobileNet/ DarkNet           & \checkmark & --         & \checkmark & \checkmark  & --        & $\vartriangle$     & $\vartriangle$ \\
LSQ+              & EfficientNet / MixNet     & \checkmark & --         & --         & \checkmark  & --         & --             & -- \\
Unified INT8      & ResNet / MobileNetV2      & \checkmark & \checkmark         & \checkmark & \checkmark  & \checkmark & \checkmark & \checkmark \\
GradScale         & ResNet                    & \checkmark & \checkmark         & \checkmark & \checkmark          & \checkmark & $\vartriangle$             & $\vartriangle$ \\
OCTAV             & ResNet / MobileNet / BERT & \checkmark & --         & \checkmark & \checkmark  & $\vartriangle$ & --          & $\vartriangle$ \\
A2Q               & General CNN               & \checkmark & --         & \checkmark & \checkmark  & --         & --             & \checkmark \\
Instance-aware DQ & ResNet / MobileNetV2      & \checkmark & \checkmark         & \checkmark & \checkmark  & --         & $\vartriangle$ & $\vartriangle$ \\
\midrule
\textbf{Coverage rate} & \textbf{(out of 12)}
& \textbf{12/12} & \textbf{5/12} & \textbf{11/12} & \textbf{10/12}
& \textbf{4/12} & \textbf{7/12} & \textbf{10/12} \\
\bottomrule
\end{tabular}%
}
\caption{Benchmark coverage of representative CNN QAT methods. Compact CNNs refer to MobileNet-, EfficientNet-, MixNet-, or similarly resource-constrained convolutional backbones. Low-bit Grad. denotes explicit low-bit gradient or low-bit training-flow evaluation. Latency/Speedup and Hardware-aware evidence are not strictly comparable because papers differ in accelerators, kernels, fixed-point assumptions, and whether real deployment is measured.}
\label{tab:cnn-coverage}
\end{table*}

\begin{table*}[!t]
\centering
\scriptsize
\setlength{\tabcolsep}{2.0pt}
\renewcommand{\arraystretch}{1.08}
\resizebox{\textwidth}{!}{%
\begin{tabular}{@{}llccccccc@{}}
\toprule
\multirow{2}{*}{\textbf{Method}} &
\multirow{2}{*}{\textbf{Backbone}} &
\multicolumn{2}{c}{\textbf{Image Classification}} &
\multicolumn{2}{c}{\textbf{ViT Architecture Coverage}} &
\multicolumn{3}{c}{\textbf{Transformer-specific Diagnostics / Deployment}} \\
\cmidrule(lr){3-4}
\cmidrule(lr){5-6}
\cmidrule(lr){7-9}
& & ImageNet-1K & Top-1 $\Delta$ & DeiT & Swin/ViT-B & Attention Dist. & Integer-only Ops & Latency/Speedup \\
\midrule
\multicolumn{9}{@{}l}{\textit{Fully quantized low-bit ViTs}} \\
\midrule
Q-ViT        & DeiT / Swin              & \checkmark & \checkmark & \checkmark & \checkmark & \checkmark & --         & $\vartriangle$ \\
OFQ          & DeiT / Swin              & \checkmark & \checkmark & \checkmark & \checkmark & \checkmark & --         & -- \\
\midrule
\multicolumn{9}{@{}l}{\textit{Integer-only and low-bit Transformer training}} \\
\midrule
I-ViT        & DeiT / Swin / ViT-B      & \checkmark & \checkmark & \checkmark & \checkmark & \checkmark & \checkmark & \checkmark \\
Xi et al.    & ViT-L/16 / DeiT-S        & \checkmark & \checkmark & \checkmark & \checkmark         & \checkmark & $\vartriangle$ & $\vartriangle$ \\
\midrule
\textbf{Coverage rate} & \textbf{(out of 4)}
& \textbf{4/4} & \textbf{4/4} & \textbf{4/4} & \textbf{4/4}
& \textbf{4/4} & \textbf{2/4} & \textbf{3/4} \\
\bottomrule
\end{tabular}%
}
\caption{Benchmark coverage of representative Vision Transformer QAT methods. Attention Dist. denotes attention-map, self-attention, query-key, or block-level distortion analysis. Integer-only Ops denotes whether nonlinear Transformer operators such as Softmax, GELU, and LayerNorm are made integer-compatible. Latency/Speedup evidence is not directly comparable because Q-ViT reports theoretical acceleration, I-ViT reports deployment-oriented speedup, and low-bit training works may focus on training data flow rather than inference kernels.}
\label{tab:vit-coverage}
\end{table*}

\begin{table*}[!t]
\centering
\scriptsize
\setlength{\tabcolsep}{2.0pt}
\renewcommand{\arraystretch}{1.08}
\resizebox{\textwidth}{!}{%
\begin{tabular}{@{}llccccccc@{}}
\toprule
\multirow{2}{*}{\textbf{Method}} &
\multirow{2}{*}{\textbf{Backbone}} &
\multicolumn{2}{c}{\textbf{NLU Benchmarks}} &
\multicolumn{2}{c}{\textbf{Compression Setting}} &
\multicolumn{3}{c}{\textbf{Deployment / Analysis}} \\
\cmidrule(lr){3-4}
\cmidrule(lr){5-6}
\cmidrule(lr){7-9}
& & GLUE & SQuAD & Ultra-low W/E & W+A / INT8 & Model Size & Latency/Speedup & Distill./Diagnostics \\
\midrule
\multicolumn{9}{@{}l}{\textit{Ultra-low-bit BERT compression}} \\
\midrule
TernaryBERT          & BERT-base              & \checkmark & \checkmark & \checkmark & \checkmark & \checkmark & --         & \checkmark \\
BinaryBERT           & BERT-base              & \checkmark & \checkmark & \checkmark & \checkmark & \checkmark & --         & \checkmark \\
\midrule
\multicolumn{9}{@{}l}{\textit{Integer-only and activation-aware Transformer encoder quantization}} \\
\midrule
Bondarenko et al.    & BERT / Transformer Enc. & \checkmark & --         & \checkmark & \checkmark & \checkmark & --         & \checkmark \\
I-BERT               & RoBERTa-base / large    & \checkmark & --         & --             & \checkmark & $\vartriangle$ & \checkmark & \checkmark \\
OCTAV                & BERT-base / large       & --         & \checkmark & $\vartriangle$     & \checkmark & --         & --         & \checkmark \\
\midrule
\textbf{Coverage rate} & \textbf{(out of 5)}
& \textbf{4/5} & \textbf{3/5} & \textbf{4/5} & \textbf{5/5}
& \textbf{4/5} & \textbf{1/5} & \textbf{5/5} \\
\bottomrule
\end{tabular}%
}
\caption{Benchmark coverage of representative BERT-family and Transformer-encoder QAT methods. Ultra-low W/E denotes binary or ternary weight and embedding quantization. W+A / INT8 denotes joint weight--activation quantization or integer-only INT8 inference. Distill./Diagnostics covers knowledge distillation, layer-wise analysis, activation-outlier analysis, or quantization-error diagnostics. Latency/Speedup remains a coverage gap because most BERT-family QAT studies emphasize accuracy and compression rather than end-to-end measured deployment.}
\label{tab:bert-coverage}
\end{table*}

\begin{table*}[!t]
\centering
\scriptsize
\setlength{\tabcolsep}{2.0pt}
\renewcommand{\arraystretch}{1.08}
\resizebox{\textwidth}{!}{%
\begin{tabular}{@{}llccccccc@{}}
\toprule
\multirow{2}{*}{\textbf{Method}} &
\multirow{2}{*}{\textbf{Backbone}} &
\multicolumn{3}{c}{\textbf{Generation Benchmarks}} &
\multicolumn{2}{c}{\textbf{Generation Metrics}} &
\multicolumn{2}{c}{\textbf{Temporal / Efficiency Evidence}} \\
\cmidrule(lr){3-5}
\cmidrule(lr){6-7}
\cmidrule(lr){8-9}
& & CIFAR-10 & LSUN & ImageNet & FID/sFID & IS/CLIP & Timestep Analysis & Sampling Cost \\
\midrule
\multicolumn{9}{@{}l}{\textit{Low-bit diffusion-model QAT / QAT-like fine-tuning}} \\
\midrule
Q-DM          & DDPM / DDIM          & \checkmark & --         & \checkmark         & \checkmark & \checkmark & \checkmark & \checkmark \\
EfficientDM   & DDIM / LDM           & \checkmark & \checkmark & \checkmark & \checkmark & \checkmark & \checkmark & \checkmark \\
\midrule
\multicolumn{9}{@{}l}{\textit{Extreme binarization for diffusion models}} \\
\midrule
BiDM          & DDIM / LDM           & \checkmark & \checkmark & --         & \checkmark & \checkmark & \checkmark & \checkmark \\
\midrule
\textbf{Coverage rate} & \textbf{(out of 3)}
& \textbf{3/3} & \textbf{2/3} & \textbf{2/3} & \textbf{3/3}
& \textbf{3/3} & \textbf{3/3} & \textbf{3/3} \\
\bottomrule
\end{tabular}%
}
\caption{Benchmark coverage of representative diffusion-model QAT methods. FID/sFID is the dominant evaluation family for quantized diffusion models, while IS/CLIP-style metrics are much less consistently reported in these representative QAT works. Timestep Analysis denotes timestep-aware quantization, timestep-wise learned step sizes, cross-timestep structure, or explicit discussion of quantization error accumulation along denoising trajectories. Sampling Cost denotes reported denoising-step, quantization-speed, storage, OPs, or efficiency evidence.}
\label{tab:diffusion-coverage}
\end{table*}

\begin{table*}[!t]
\centering
\scriptsize
\setlength{\tabcolsep}{2.0pt}
\renewcommand{\arraystretch}{1.08}
\resizebox{\textwidth}{!}{%
\begin{tabular}{@{}llcccccc@{}}
\toprule
\multirow{2}{*}{\textbf{Method}} &
\multirow{2}{*}{\textbf{Backbone}} &
\multicolumn{2}{c}{\textbf{Graph Benchmarks}} &
\multicolumn{2}{c}{\textbf{Quantization Setting}} &
\multicolumn{2}{c}{\textbf{Graph-specific Evaluation}} \\
\cmidrule(lr){3-4}
\cmidrule(lr){5-6}
\cmidrule(lr){7-8}
& & Citation Graphs & \shortstack{Inductive /\\Unseen Graphs} & INT8 & INT4 & Degree Analysis & \shortstack{Unseen Graph\\Generalization} \\
\midrule
\multicolumn{8}{@{}l}{\textit{GNN-targeted quantization-aware training}} \\
\midrule
Degree-Quant & GCN / GAT / GIN
& \checkmark
& \checkmark
& \checkmark
& $\vartriangle$
& \checkmark
& \checkmark \\
\midrule
\textbf{Coverage rate} & \textbf{(out of 1)}
& \textbf{1/1}
& \textbf{1/1}
& \textbf{1/1}
& \textbf{$\vartriangle$}
& \textbf{1/1}
& \textbf{1/1} \\
\bottomrule
\end{tabular}%
}
\caption{Benchmark coverage of representative GNN QAT methods. Degree-Quant is the primary GNN-targeted QAT method in this taxonomy. Citation Graphs denotes transductive citation-network evaluation, while Inductive / Unseen Graphs denotes graph-classification, graph-regression, or unseen-graph evaluation. INT4 is marked with $\triangle$ because it is reported only for GIN on REDDIT-BINARY, rather than across all evaluated backbones and datasets. Degree Analysis captures the central motivation of Degree-Quant: quantization error depends on graph degree and aggregation statistics. The coverage rate is shown out of one representative method, rather than used to imply that GNN QAT is a mature or densely covered area.}
\label{tab:gnn-coverage}
\end{table*}

\subsection{Extended Structural Evaluation}
\label{subsec:structural-eval}

Structural evaluation is needed to identify where quantization error enters and how it propagates. Tensor distortion can be measured with normalized mean squared error:
\[
\mathrm{nMSE}(X)=
\frac{\|X-Q(X)\|_F^2}{\|X\|_F^2}.
\]
This metric is often complemented by cosine similarity, clipping ratio, saturation ratio, and layer-wise error profiles \citep{choi2018pact,esser2020learned,bhalgat2020lsq+,jain2020trained,zhao2020linear}. Reporting per-layer curves is preferable to a single aggregate number because quantization error is usually concentrated in a few sensitive modules \citep{bondarenko2021understanding,li2022qvit,chen2025efficientqat}.

For Transformers, tensor-level distortion should be paired with block-output drift and logit divergence, since small perturbations can be amplified by attention, residual connections, normalization, and nonlinear functions \citep{bondarenko2021understanding,li2022qvit,kim2021bert,li2023vit}. Useful diagnostics include block-output MSE, hidden-state cosine similarity, KL divergence between full-precision and quantized logits, and the change in cross-entropy or perplexity on the same evaluation batches \citep{Hinton2015DistillingTK,kim-etal-2022-understanding,liu2024llm}. These metrics connect local quantization noise to functional degradation.

Training stability is a QAT-specific structural criterion. Since rounding is non-differentiable, most QAT methods rely on STE-style surrogate gradients \citep{bengio2013estimating,jacob2018quantization,esser2020learned}. We therefore recommend reporting gradient cosine similarity, sign agreement, gradient-norm drift, and the fraction of weights whose quantized values repeatedly flip across adjacent updates \citep{zhu2020towards,sun2020ultra,nagel2022overcoming,liu2023oscillation}. These diagnostics are especially important below 3 bits, where gradient mismatch and threshold oscillation often dominate final performance \citep{yin2019understanding,nagel2022overcoming,liu2023oscillation}.

Hardware feasibility should also be evaluated structurally. For integer matrix multiplication, the accumulator range depends on both operand ranges and the reduction dimension \citep{jacob2018quantization,colbert2023a2q}. A conservative bound is
\[
\left|\sum_{i=1}^{n} q_i^w q_i^a\right|
\leq n \cdot q_{\max}^{w}\cdot q_{\max}^{a}.
\]
Thus, the accumulator bit-width may be much larger than the weight or activation bit-width. Hardware-aware QAT should report analytical safety margins and empirical overflow or saturation rates under representative inputs, as exemplified by accumulator-aware QAT \citep{colbert2023a2q}.

\subsection{Extended Reasoning Evaluation}
\label{subsec:reasoning-eval}

For LLM QAT, reasoning should be evaluated separately from language modeling. Perplexity is useful for measuring average next-token fidelity, but reasoning requires maintaining intermediate states across long generated sequences, as emphasized by chain-of-thought prompting studies \citep{wei2022chain}. Quantization noise in weights, activations, or KV cache can accumulate across layers and decoding steps \citep{liu2024llm,bondarenko2024low}, making reasoning degradation visible even when perplexity changes are small.

A minimal reasoning protocol should include both answer-only and rationale-generating prompts. GSM8K or MATH-style benchmarks measure mathematical reasoning \citep{cobbe2021training,hendrycks2021math}; BBH or MMLU-style benchmarks test multi-step knowledge reasoning \citep{suzgun2023challenging,hendrycks2020mmlu}; and HumanEval or MBPP-style benchmarks evaluate code generation \citep{chen2021evaluating,austin2021program}. The prompt template, number of shots, decoding temperature, maximum generation length, answer-extraction rule, and invalid-output handling should be fixed and reported. We recommend reporting final-answer accuracy, invalid-output rate, and generation length.

When possible, LLM QAT papers should compare W-only, W+A, W+A+KV, and KV-only settings under the same prompting and decoding configuration \citep{liu2024llm,bondarenko2024low,chen2025efficientqat}. This isolates whether reasoning loss is caused mainly by weight discretization, activation outliers, KV-cache accumulation, or their interaction. Such target-specific evaluation is particularly important for long-context generation, where KV-cache errors may remain small at each step but accumulate over the whole decoding trajectory \citep{liu2024llm,bondarenko2024low}.

\subsection{Extended Efficiency and Deployment Evaluation}
\label{subsec:efficiency-eval}

Efficiency evaluation should distinguish model compression, training feasibility, and realized inference speed \citep{Nagel2021AWP,jacob2018quantization}. Model footprint should include quantized weights, scales, zero-points, preserved full-precision tensors, adapters, and KV-cache storage when applicable. This is especially important for group-wise QAT and Q-PEFT, where scale tensors or adapter parameters may occupy non-negligible memory \citep{hu2022lora,xu2023qa,li2023loftq,bondarenko2024low,chen2025efficientqat}. A reported compression ratio should therefore reflect actual storage rather than nominal weight bit-width alone.

Training efficiency should be reported independently of inference efficiency. QAT often maintains full-precision master weights, optimizer states, activation checkpoints, and sometimes full-precision teacher models for distillation \citep{jacob2018quantization,Nagel2021AWP,Hinton2015DistillingTK,liu2024llm}. Hence, practical feasibility depends on peak memory, wall-clock time, number of steps or tokens, batch size, sequence length, optimizer precision, gradient checkpointing, and hardware type \citep{liu2024llm,YuanSD24,bondarenko2024low,chen2025efficientqat}. For LLM QAT, whether a method fits on a single commodity GPU can be more informative than its nominal FLOP count \citep{bondarenko2024low,chen2025efficientqat,dremov2025compute}.

Inference efficiency should be measured on the intended deployment path rather than inferred from bit-width \citep{jacob2018quantization,kim2021bert,li2023vit}. For autoregressive LLMs, prefill latency and decoding throughput should be reported separately, since weight quantization and KV-cache quantization affect these stages differently \citep{liu2024llm,bondarenko2024low}. Reports should specify prompt length, generated length, context length, batch size, KV-cache precision, kernel backend, compiler stack, and whether any operators fall back to floating point. For vision and encoder models, latency should specify input resolution, batch size, and whether Softmax, GELU, and LayerNorm are executed in integer arithmetic \citep{kim2021bert,li2023vit}.

We distinguish four deployment levels. Level 1 is fake quantization, where low-bit effects are simulated in floating point \citep{jacob2018quantization,Nagel2021AWP}. Level 2 is simulated integer arithmetic, where integer ranges and rounding are modeled but kernels are not deployed \citep{colbert2023a2q}. Level 3 is partial integer deployment, where matrix multiplications use integer kernels but nonlinear functions or normalization remain in floating point \citep{kim2021bert,li2023vit}. Level 4 is end-to-end integer deployment, where linear layers, nonlinear functions, normalization, and accumulators are compatible with the target integer backend \citep{kim2021bert,li2023vit}. QAT papers should state which level their evaluation supports.

\subsection{Case study of the Minimal Reporting Checklist}
\label{app:case_study}
We apply the Minimal Reporting Checklist to LLM-QAT \citep{liu2024llm}. The study provides a relatively complete quantization configuration, specifying the quantization targets (weights, activations, and KV cache), numerical scheme, and granularity, including per-channel weight and per-token activation/KV-cache quantization. It also evaluates multiple W--A--KV precision configurations against full-precision baselines. However, absolute performance deltas from the matched full-precision baseline are not systematically reported, and explicit structural diagnostics such as layer-wise distortion, gradient mismatch, or overflow rate are absent. In terms of efficiency, LLM-QAT reports model size, operation counts, and GPU costs for data generation and QAT, although direct kernel-level latency or throughput measurements are not provided. For LLM evaluation, the study includes 5-shot MMLU in addition to perplexity and commonsense reasoning benchmarks.

\section{Supplementary Tables}
\label{app:supple}
Appendix \ref{app:supple} provides a compact table-based supplement to the taxonomy in §\ref{sec:taxonomy_target} and the evaluation synthesis in §\ref{sec:eval}. Table \ref{tab:qat_taxonomy} gives the method-level view corresponding to Figure \ref{fig:qat_taxonomy}, Tables \ref{tab:cross_target_comparison}–\ref{tab:int_fp_comparison} summarize cross-target properties and numerical-format trade-offs, and Tables \ref{tab:imagenet-cnn}–\ref{tab:instruction_tuned_results} collect representative quantitative evidence across CNNs, ViTs, BERT-style encoders, LLMs, and diffusion models. Because the original studies differ in quantization targets, training budgets, evaluation protocols, and deployment assumptions, the quantitative tables should be interpreted primarily through within-paper deltas.

\paragraph{Notation and reporting convention.} W, A, G, KV, E, and Opt denote weights, activations, gradients, key-value cache, embeddings, and optimizer states, respectively. The notation g64/g128 indicates group size, while NF2/NF4 denotes 2-/4-bit NormalFloat formats. Values in parentheses denote differences from the corresponding full-precision baseline reported in the original paper, while “–” indicates unavailable results. 

Emerging low-bit formats span several related but distinct settings. 
We treat FP8 QAT \citep{kuzmin2022fp8} and learnable non-uniform quantizers \citep{gongyo2024learning} as part of the core QAT discussion, while microscaling formats \citep{rouhani2023microscaling} and fully quantized FP4 training \citep{Chmiel2025FP4} are considered adjacent evidence. 
These neighboring areas remain relevant because they expose format-specific constraints on dynamic range, scaling, rounding, accumulation, and hardware execution \citep{kuzmin2022fp8, micikevicius2022fp8,rouhani2023microscaling, Chmiel2025FP4}.

\begin{table*}[!t]
  \centering
  \scriptsize
  \setlength{\tabcolsep}{3pt}
  \renewcommand{\arraystretch}{1.08}

  \begin{tabularx}{\textwidth}{@{}
    >{\raggedright\arraybackslash}p{2.6cm}
    >{\raggedright\arraybackslash}p{1.9cm}
    >{\centering\arraybackslash}p{1.3cm}
    >{\raggedright\arraybackslash}X
    >{\raggedright\arraybackslash}p{2.2cm}
    >{\raggedright\arraybackslash}p{2.0cm}
  @{}}
    \toprule
    \textbf{Approach} & \textbf{Venue} & \textbf{Target} &
    \textbf{Bits} & \textbf{Gran.} & \textbf{Model} \\
    \midrule

    TWN/TTQ       & 2016 arXiv              & W        & W2                         & Tensor                   & CNN / General-CV \\
TernaryBERT   & 2020 EMNLP              & W        & W2E2A8                     & Layer, Row               & BERT \\
BinaryBERT    & 2021 ACL                & W        & W1E1A8, W1E1A4             & Layer, Tensor            & BERT \\
\citet{nagel2022overcoming}  & 2022 ICML               & W(+A)    & W3A3, W4A4                 & Tensor                   & CNN / General \\
BitNet        & 2023 arXiv              & W        & W1A8                       & Tensor                   & LLM \\
A2Q           & 2023 ICCV               & W        & W2, W3, W4, W5             & Channel                  & General / CNN \\
OFQ           & 2023 ICML               & W(+A)    & W2A2, W3A3, W4A4           & Row                      & ViT \\
BitNet b1.58 (\citeauthor{ma2024era}) & 2024 arXiv   & W        & W1.58A8                    & Tensor                   & LLM \\
BitNet b1.58 Reloaded    & 2024 DeLTA   & W        & W1.58                      & Tensor                   & Small LM / Vision \\
DL-QAT        & 2024 EMNLP Industry     & W        & W3, W4                     & Group                    & LLM \\
QEFT          & 2024 EMNLP Findings     & W        & W3, W4                     & Group                    & LLM \\
PB-LLM        & 2024 ICLR               & W        & partial W1 + salient FP    & Column / Salient weights & LLM \\
QA-LoRA       & 2024 ICLR               & W        & W2, W3, W4                 & Group                    & LLM \\
LoftQ         & 2024 ICLR               & W        & W2, W4; NF2, NF4           & Group / Block            & LLM \\
LR-QAT        & 2024 arXiv              & W        & W3, W4                     & Channel, Group           & LLM \\
EfficientQAT  & 2025 ACL                & W        & W2g64, W2g128, W3g128, W4g128 & Group                 & LLM \\
L4Q           & 2025 ACL                & W        & W3, W4                     & Group                    & LLM \\
CQPT          & 2025 ACL Findings       & W        & FP16 $\rightarrow$ W1.58    & Layer, Token             & BitNet LM \\
ParetoQ       & 2025 NeurIPS            & W        & W1, W1.58, W2, W3, W4      & /                        & LLM \\

PACT          & 2018 arXiv              & A        & W2A2, W3A3, W4A4           & Layer                    & CNN / General-CV \\
LSQ+          & 2020 CVPRW              & A        & W2A2, W3A3, W4A4           & Layer                    & Efficient CNNs \\
BinaryDuo     & 2020 ICLR               & A        & W1A1; coupled A1 proxy     & /                        & BNN \\

XNOR-Net      & 2016 ECCV               & W + A    & BWN: W1; XNOR: W1A1        & Tensor                   & CNN / BNN \\
DSQ           & 2019 ICCV               & W + A    & W1A1, W2A2, W3A3, W4A4     & Layer                    & CNN / General-CV \\
LSQ           & 2020 ICLR               & W + A    & W2A2, W3A3, W4A4           & Layer                    & CNN / General-CNN \\
LLSQ          & 2020 ICLR               & W + A    & W3A3, W4A4                 & Channel, Layer           & CNN \\
\citet{jain2020trained}   & 2020 MLSys              & W + A    & W8A8, W4A4                 & Tensor                   & CNN \\
\citet{bondarenko2021understanding} & 2021 EMNLP          & W + A    & W8A8; low-bit W/E ablations & Tensor / Embedding-group & BERT / Transformer encoder \\
I-BERT        & 2021 ICML               & W + A    & W8A8                       & /                        & BERT / RoBERTa \\
Degree-Quant  & 2021 ICLR               & W + A    & INT8, INT4                 & Tensor                   & GNN \\
\citet{liu2022instance}    & 2022 CVPR               & W + A    & dynamic W/A 2--6; W4A4     & Layer, Instance          & CNN / Image classification \\
OCTAV         & 2022 ICML               & W + A    & W4A4, W5A5, W6A6           & Tensor / Vector          & CNN + BERT \\
\citet{kuzmin2022fp8} & 2022 NeurIPS    & W + A    & W8A8 (FP8)     & Tensor, Channel  & CNN + BERT \\
Q-ViT         & 2022 NeurIPS            & W + A    & W2A2, W3A3, W4A4           & /                        & ViT \\
I-ViT         & 2023 ICCV               & W + A    & W8A8                       & /                        & ViT \\
nu-LSQ        & 2024 ACCV               & W + A    & W2A2, W3A3, W4A4           & Layer / Non-uniform steps & CNN / ViT \\
EfficientDM   & 2024 ICLR Spotlight     & W + A    & W4A8, W4A4                 & Channel, Layer           & Diffusion Model \\
BiDM          & 2024 NeurIPS            & W + A    & W1A1; W1A8, W2A8           & Layer                    & Diffusion Model \\

DoReFa-Net    & 2016 arXiv              & W + A + G & W1A2G6                    & Tensor                   & CNN \\
\citet{zhu2020towards}    & 2020 CVPR               & W + A + G & W8A8G8                    & Layer                    & CNN \\
GradScale     & 2020 NeurIPS            & W + A + G & W4A4G4                    & Layer                    & General \\
\citet{xi2023training}     & 2023 NeurIPS            & W + A + G & W4A4G4 / INT4 training    & Token / Structure-aware  & Transformer \\
\citet{chitsaz-etal-2024-exploring} & 2024 EMNLP Findings     & W + A + G + Opt & 4, 8                & Tensor, Channel, Token   & Transformer LM \\

LLM-QAT       & 2024 ACL Findings       & W + A + KV & W4A8KV4, W4A8KV8         & Token                    & LLM \\
    \bottomrule
  \end{tabularx}
 \caption{Taxonomy of quantization-aware training (QAT) frameworks.}
  \label{tab:qat_taxonomy}
\end{table*}

\begin{table*}[t]
\centering
\footnotesize
\setlength{\tabcolsep}{3pt}
\renewcommand{\arraystretch}{1.05}

\begin{tabularx}{\textwidth}{
    >{\centering\arraybackslash}p{0.055\textwidth}
    >{\raggedright\arraybackslash}X
    >{\raggedright\arraybackslash}X
    >{\raggedright\arraybackslash}X
    >{\raggedright\arraybackslash}X
}
\toprule
\textbf{Target} &
\textbf{Target properties} &
\textbf{Dominant failure modes} &
\textbf{Error propagation} &
\textbf{Typical adaptation} \\
\midrule

\textbf{W} &
Persistent; relatively stationary; heterogeneous sensitivity. &
Salient-weight distortion; grid-point oscillation; limited ultra-low-bit capacity. &
Persistent errors reused across layers and inputs. &
Per-channel/group scaling; sensitivity-aware precision; stabilization. \\
\midrule

\textbf{A} &
Input-, token-, and layer-dependent; structured outliers; large dynamic range. &
Clipping/saturation; outlier-dominated scales; range mismatch. &
Input-conditioned errors alter downstream distributions. &
Fine-grained scaling; learnable clipping or step sizes. \\
\midrule

\textbf{W+A} &
Coupled quantization of both linear operands. &
Coupled W/A distortion; accumulator overflow; integerization mismatch. &
$\Delta WA+W\Delta A+\Delta W\Delta A$ jointly perturb representations. &
Separate W/A granularity; accumulator- and operator-aware design. \\
\midrule

\textbf{KV} &
Runtime-generated; sequence-persistent; K/V heterogeneous. &
K/V mismatch; cache outliers; unsuitable shared granularity. &
Reused cache errors affect future attention across decoding steps. &
K/V-specific token-, channel-, or group-wise quantization. \\
\midrule

\textbf{G} &
Training-only; highly dynamic; structurally non-uniform. &
Under/overflow; small-gradient loss; direction distortion; instability. &
Perturbed updates alter subsequent optimization trajectories. &
Dynamic scaling; selective precision; direction-/structure-aware quantization. \\

\bottomrule
\end{tabularx}

\caption{Cross-target comparison of target properties, dominant failure
modes, error propagation, and adaptation strategies.}
\label{tab:cross_target_comparison}
\end{table*}

\begin{table*}[t]
\centering
\scriptsize
\setlength{\tabcolsep}{3.5pt}
\renewcommand{\arraystretch}{1.15}

\begin{tabularx}{\textwidth}{
    >{\raggedright\arraybackslash}p{0.16\textwidth}
    >{\raggedright\arraybackslash}X
    >{\raggedright\arraybackslash}X
    >{\raggedright\arraybackslash}p{0.23\textwidth}
}
\toprule

\textbf{Dimension} &
\textbf{INT8/INT4-based QAT} &
\textbf{FP8/FP4-based QAT or training} &
\textbf{Target-centric implication} \\

\midrule

\textbf{Representation and range} &
Uniform grid, fixed absolute step, and scale-defined range\citep{jacob2018quantization,esser2020learned}. &
Exponent--mantissa grid, magnitude-dependent spacing, and wider range\citep{micikevicius2022fp8,kuzmin2022fp8}. &
INT favors bounded distributions; FP favors wide dynamic ranges. \\

\midrule

\textbf{Quantization error} &
In-range rounding plus out-of-range clipping\citep{jacob2018quantization,esser2020learned}. &
Relative rounding plus underflow/overflow; FP4 has very limited mantissa precision\citep{kuzmin2022fp8,wang2025optimizing}. &
Error type matters more than nominal bit-width alone. \\

\midrule

\textbf{Scaling and granularity} &
Supports tensor-, channel-, group-, or block-wise scales\citep{esser2020learned,bondarenko2021understanding}. &
Still scale-dependent; FP4 commonly requires block/microscaling\citep{rouhani2023microscaling,Chmiel2025FP4}. &
Report scale type, granularity, and block size. \\

\midrule

\textbf{Activation outliers} &
Outliers enlarge the scale and waste integer levels\citep{bondarenko2021understanding,xi2023training}. &
The exponent reduces clipping, although FP4 remains sensitive to outliers\citep{kuzmin2022fp8,wang2025optimizing}. &
FP mitigates but does not eliminate the outlier problem. \\

\midrule

\textbf{Effect of QAT} &
Learned scales, clipping, and parameters reshape distributions toward the integer grid\citep{kuzmin2022fp8}. &
The FP8 advantage observed in PTQ can narrow substantially after QAT\citep{kuzmin2022fp8}. &
Neither FP nor INT is universally superior under QAT. \\

\midrule

\textbf{Training stability} &
INT8 is relatively mature; INT4 W/A/G quantization requires specialized clipping and gradient treatment\citep{jacob2018quantization,xi2023training}. &
FP8 better accommodates changing ranges; FP4 requires fine-grained scaling, careful rounding, and higher-precision accumulation\citep{Chmiel2025FP4,fishman2025scaling}. &
Forward, backward, update, and accumulator precision must be distinguished. \\

\midrule

\textbf{Hardware support} &
INT8 kernels are mature; INT4 support is more target- and backend-dependent\citep{jacob2018quantization,xi2023training}. &
FP8 support is increasing; FP4/MX support remains format- and hardware-specific\citep{Chmiel2025FP4,fishman2025scaling}. &
Fake-quantized accuracy does not guarantee deployment gains. \\

\midrule

\textbf{Typical use} &
INT8 is common for W+A inference; INT4 is often weight-centric\citep{jacob2018quantization,liu2024llm}. &
FP8 supports W/A/G low-precision training; FP4 W/A/G training is emerging\citep{micikevicius2022fp8,Chmiel2025FP4}. &
Format selection should follow the quantization target. \\

\midrule

\textbf{W/A/KV/G interaction} &
Stable weights suit group-wise INT; A/KV/G require more dynamic scaling\citep{xi2023training,liu2024llm,liu2024kivi}. &
FP range benefits dynamic activations and gradients but does not prevent KV-cache error accumulation\citep{micikevicius2022fp8,Chmiel2025FP4}. &
Hybrid, target-specific INT/FP policies may be preferable. \\

\bottomrule
\end{tabularx}
\caption{Comparison of INT8/INT4-based and FP8/FP4-based quantization-aware training and low-precision training.}
\label{tab:int_fp_comparison}
\end{table*}

\begin{table}[!tbp]
\centering
\resizebox{\columnwidth}{!}{
\begin{tabular}{llcc}
\toprule
Method & Model & Bits & Top-1 (\%) ($\Delta$) \\
\midrule
\citeauthor{zhu2020towards}      & ResNet-18    & W8A8G8 & 69.67 ($-0.63$) \\
\citeauthor{zhu2020towards}      & ResNet-50    & W8A8G8 & 76.34 ($-0.26$) \\
\citeauthor{zhu2020towards}      & MobileNetV2  & W8A8G8 & 71.20 ($-1.19$) \\
LLSQ            & ResNet-18    & W4A4   & 69.84 ($+0.08$) \\
LLSQ            & ResNet-18    & W3A3   & 68.08 ($-1.68$) \\
LLSQ            & MobileNetV2  & W4A4   & 67.37 ($-4.43$) \\
TQT             & MobileNet-V2 & W8A8   & 71.80 ($+0.10$) \\
OCTAV           & ResNet-50    & W4A4   & 76.21 ($+0.14$) \\
OCTAV           & ResNet-18    & W4A4   & 69.90 ($-0.22$) \\
OCTAV           & MobileNet-V2 & W4A4   & 71.23 ($-0.48$) \\
LSQ+            & ResNet-18    & W2A2   & 66.80 ($-3.30$) \\
LSQ+            & ResNet-18    & W3A3   & 69.30 ($-0.80$) \\
LSQ+            & ResNet-18    & W4A4   & 70.80 ($+0.70$) \\
nu-LSQ-A        & ResNet-18    & W2A2   & 64.89 ($-4.87$) \\
DSQ             & ResNet-18    & W2A2   & 65.17 ($-4.73$) \\
DSQ             & ResNet-18    & W4A4   & 69.56 ($-0.34$) \\
LSQ             & ResNet-18    & W2A2   & 67.60 ($-2.90$) \\
LSQ             & ResNet-18    & W4A4   & 71.10 ($+0.60$) \\
DoReFa-Net      & AlexNet      & W1A2G6 & 46.10 ($-9.80$) \\
DoReFa-Net      & AlexNet      & W1A4G6 & 48.20 ($-7.70$) \\
\citeauthor{liu2022instance} & ResNet-50    & $\sim$4MP/$\sim$4MP & 76.94 ($-0.02$) \\
\citeauthor{liu2022instance} & MobileNetV2  & $\sim$4MP/$\sim$4MP & 72.05 (--) \\
\bottomrule
\end{tabular}
}

\caption{ImageNet Top-1 Accuracy (\%) for Representative CNN Quantization-Aware Training Methods. For \citet{liu2022instance}, $\sim$4MP/$\sim$4MP denotes the approximately 4-bit mixed/dynamic precision setting.}
\label{tab:imagenet-cnn}
\end{table}

\begin{table}[!tbp]
\centering
\resizebox{\columnwidth}{!}{
\begin{tabular}{llcc}
\toprule
Method & Backbone & Bits & Top-1 \% \\
\midrule
Q-ViT     & DeiT-S   & W4A4 & 80.90 ($+1.00$) \\
Q-ViT     & DeiT-S   & W2A2 & 72.10 ($-7.80$) \\
Q-ViT     & Swin-T   & W4A4 & 82.50 ($+1.30$) \\
Q-ViT     & Swin-T   & W2A2 & 74.70 ($-6.50$) \\
I-ViT     & DeiT-S   & W8A8 & 80.12 ($+0.27$) \\
I-ViT     & Swin-T   & W8A8 & 81.50 ($+0.15$) \\
I-ViT     & ViT-B    & W8A8 & 84.76 ($+0.23$) \\
OFQ       & DeiT-S   & W2A2 & 75.72 ($-4.18$) \\
OFQ       & Swin-T   & W2A2 & 78.52 ($-2.68$) \\
OFQ       & DeiT-S   & W4A4 & 81.10 ($+1.20$) \\
OFQ       & Swin-T   & W4A4 & 81.88 ($+0.68$) \\
\citeauthor{xi2023training} & ViT-L/16 & W4A4 & 82.61 ($-1.94$) \\
\citeauthor{xi2023training} & DeiT-S   & W4A4 & 69.18 ($-3.92$) \\
\bottomrule
\end{tabular}
}

\caption{ImageNet Top-1 Accuracy (\%) --- Vision Transformer Architectures. I-ViT INT8 results are written as W8A8 for consistency, and \citeauthor{xi2023training} report 4-bit fully quantized training rather than standard QAT-based inference quantization.}
\label{tab:imagenet-vit}
\end{table}

\begin{table}[!tbp]
\centering
\resizebox{\columnwidth}{!}{
\begin{tabular}{llccc}
\toprule
Method & Base Model & Benchmark & Bits & FID $\downarrow$ \\
\midrule
EfficientDM & DDIM  & CIFAR-10         & W4A4 & 10.48 ($+6.34$) \\
EfficientDM & LDM-4 & LSUN-Bedrooms    & W4A4 & 10.60 ($+7.17$) \\
EfficientDM & LDM-8 & LSUN-Churches    & W4A4 & 14.34 ($+10.26$) \\
BiDM        & DDIM  & CIFAR-10         & W1A1 & 81.65 ($+76.11$) \\
BiDM        & LDM-4 & LSUN-Bedrooms    & W1A1 & 22.74 ($+19.75$) \\
BiDM        & LDM-8 & LSUN-Churches    & W1A1 & 29.70 ($+25.34$) \\
\bottomrule
\end{tabular}
}

\caption{FID Results for Quantized Diffusion Models.}
\label{tab:dm-quantization}
\end{table}

\begin{table}[!tbp]
\centering
\scriptsize
\setlength{\tabcolsep}{2.2pt}
\resizebox{\columnwidth}{!}{%
\begin{tabular}{@{}lllcc@{}}
\toprule
Method & Backbone & Bits & MNLI-m (\%) & SQuAD F1 (\%) \\
\midrule
TernaryBERT & BERT-base & W2E2A8 & 83.3($-1.2$) & 87.4($-1.3$) \\
BinaryBERT & BERT-base & W1E1A8 & 84.2($-0.4$) & 88.3($-0.2$) \\
BinaryBERT & BERT-base & W1E1A4 & 83.9($-0.7$) & 87.2($-1.3$) \\
I-BERT & RoBERTa-Base & W8A8 & 87.5($-0.3$) & -- \\
I-BERT & RoBERTa-Large & W8A8 & 90.4($+0.4$) & -- \\
OCTAV Dynamic & BERT-Base & W4 & -- & 84.51($-3.73$) \\
OCTAV Dynamic & BERT-Base & W8 & -- & 88.34($+0.10$) \\
OCTAV Dynamic & BERT-Large & W4 & -- & 87.09($-3.91$) \\
OCTAV Dynamic & BERT-Large & W8 & -- & 90.78($-0.22$) \\
\bottomrule
\end{tabular}%
}
\caption{GLUE MNLI-m and SQuAD v1.1 Results for BERT-family Models. I-BERT reports INT8 integer-only inference results. Bits of OCTAV  denotes the uniform quantization bit-width used in its BERT fine-tuning experiments.}
\label{tab:nlp-bert}
\end{table}

\begin{table}[!tbp]
\centering
\scriptsize
\setlength{\tabcolsep}{2.2pt}
\resizebox{\columnwidth}{!}{%
\begin{tabular}{@{}lllcc@{}}
\toprule
Method & Base Model & Bits & PPL $\downarrow$ & Avg. zero-shot \\
\midrule
LLM-QAT & LLaMA-1-7B  & W4A8KV8   & 11.2($+0.8$)  & 64.8($-1.4$) \\
LLM-QAT & LLaMA-1-7B  & W4A8KV4   & 11.6($+1.2$)  & 63.0($-3.2$) \\
LLM-QAT & LLaMA-1-13B & W4A8KV8   & 10.0($+0.3$)  & 66.8($-1.2$) \\
LLM-QAT & LLaMA-1-13B & W4A8KV4   & 10.2($+0.5$)  & 63.4($-4.6$) \\

LR-QAT & LLaMA-1-7B  & W4 g128   & 5.75($+0.07$) & 69.15($-0.53$) \\
LR-QAT & LLaMA-2-7B  & W4 g128   & 5.59($+0.12$) & 69.88($-0.59$) \\
LR-QAT & LLaMA-2-13B & W4 g128   & 4.97($+0.09$) & 72.91($-0.27$) \\

EfficientQAT & LLaMA-2-7B  & W3g128 & 5.81($+0.34$) & 64.02($-0.84$) \\
EfficientQAT & LLaMA-2-7B  & W2g64  & 6.86($+1.39$) & 60.14($-4.72$) \\
EfficientQAT & LLaMA-2-13B & W3g128 & 5.12($+0.24$) & 67.28($-0.53$) \\
EfficientQAT & LLaMA-2-13B & W2g64  & 5.96($+1.08$) & 64.48($-3.33$) \\

DL-QAT & LLaMA-1-7B & W4A8KV8    & 6.7($+1.02$)  & 69.4($-1.1$) \\
DL-QAT & LLaMA-2-7B & W4A16KV16  & 6.3($+0.83$)  & 69.3($-1.2$) \\

PB-LLM & LLaMA-1-7B & W1+30\%W8  & --             & 66.9($-1.8$) \\
PB-LLM & LLaMA-1-7B & W1+10\%W8  & --             & 60.6($-8.1$) \\

QA-LoRA & LLaMA-1-7B & W4 g32     & --             & 61.8($+3.5$) \\
QA-LoRA & LLaMA-1-7B & W2 g32     & --             & 53.7($-4.6$) \\

LoftQ & LLaMA-2-7B  & W4 NF4      & 5.24($+0.16$) & -- \\
LoftQ & LLaMA-2-7B  & W2 NF2      & 7.85($+2.77$) & -- \\
LoftQ & LLaMA-2-13B & W4 NF4      & 5.16($+0.04$) & -- \\
LoftQ & LLaMA-2-13B & W2 NF2      & 7.69($+2.57$) & -- \\

L4Q & LLaMA-1-7B  & W4 g128      & --             & 62.7($+1.0$) \\
L4Q & LLaMA-1-13B & W4 g128      & --             & 64.5($+0.7$) \\
L4Q & LLaMA-2-7B  & W4 g128      & --             & 63.6($+1.7$) \\
L4Q & LLaMA-2-13B & W4 g128      & --             & 65.8($+0.8$) \\
\bottomrule
\end{tabular}%
}
\caption{WikiText-2 perplexity and zero-shot accuracy for base LLMs. The baseline of LoftQ is the 16-bit LoRA model. Avg. zero-shot scores are paper-specific aggregates and may differ in task composition and evaluation protocol across studies.}
\label{tab:llm_results}
\end{table}

\begin{table}[!tbp]
\centering
\scriptsize
\setlength{\tabcolsep}{2.5pt}
\resizebox{\columnwidth}{!}{%
\begin{tabular}{@{}lllcc@{}}
\toprule
\textbf{Method} & \textbf{Base Model} & \textbf{Bits} 
& \textbf{MMLU} $\uparrow$ & \textbf{Avg.} $\uparrow$ \\
\midrule
QA-LoRA & LLaMA-1-7B & W4 & 39.4($+4.8$) & 61.8($+3.5$) \\
QA-LoRA & LLaMA-1-7B & W3 & 37.4($+2.8$) & 60.1($+1.8$) \\
QA-LoRA & LLaMA-1-7B & W2 & 27.5($-7.1$) & 53.7($-4.6$) \\
DL-QAT & LLaMA-1-7B & W4 & 39.9($+5.3$) & 63.4($+5.1$) \\
DL-QAT & LLaMA-1-7B & W3 & 33.9($-0.7$) & 61.1($+2.8$) \\
L4Q & LLaMA-1-7B & W4 & 35.7($+0.6$) & 62.7($+1.0$) \\
L4Q & LLaMA-1-7B & W3 & 31.8($-3.3$) & 61.2($-0.5$) \\
\bottomrule
\end{tabular}%
}
\caption{Instruction-tuned Quantized LLMs on MMLU and Commonsense Benchmarks. MMLU denotes 5-shot MMLU accuracy. Avg. denotes the reported zero-shot commonsense QA average}
\label{tab:instruction_tuned_results}
\end{table}

\end{document}